# Probabilistic Inference from Arbitrary Uncertainty
# using Mixtures of Factorized Generalized Gaussians

**Alberto Ruiz**                                                     ARUIZ@DIF.UM.ES
**Pedro E. López-de-Teruel**                                      PEDROE@DITEC.UM.ES
**M. Carmen Garrido**                                            MGARRIDO@DIF.UM.ES
*Universidad de Murcia, Facultad de Informática,*
*Campus de Espinardo, 30100, Murcia, Spain*

## Abstract

This paper presents a general and efficient framework for probabilistic inference and learning from arbitrary uncertain information. It exploits the calculation properties of finite mixture models, conjugate families and factorization. Both the joint probability density of the variables and the likelihood function of the (objective or subjective) observation are approximated by a special mixture model, in such a way that any desired conditional distribution can be directly obtained without numerical integration. We have developed an extended version of the expectation maximization (EM) algorithm to estimate the parameters of mixture models from uncertain training examples (indirect observations). As a consequence, any piece of exact or uncertain information about both input and output values is consistently handled in the inference and learning stages. This ability, extremely useful in certain situations, is not found in most alternative methods. The proposed framework is formally justified from standard probabilistic principles and illustrative examples are provided in the fields of nonparametric pattern classification, nonlinear regression and pattern completion. Finally, experiments on a real application and comparative results over standard databases provide empirical evidence of the utility of the method in a wide range of applications.

## 1. Introduction

The estimation of unknown magnitudes from available information, in the form of sensor measurements or subjective judgments, is a central problem in many fields of science and engineering. To solve this task, the domain must be accurately described by a model able to support the desired range of inferences. When satisfactory models cannot be derived from first principles, approximations must be obtained from empirical data in a learning stage.

Consider a domain $Z$ composed by a collection of objects $z = (z^1, z^2, ..., z^n)$, represented by vectors of $n$ attributes. Given some partial knowledge $S$ (expressed in a general form explained later) about a certain object $z$, we are interested in computing a good estimate $\hat{z}(S)$, close to the true $z$. We allow heterogeneous descriptions; any attribute $z^i$ may be continuous, discrete, or symbolic valued, including mixed types. If there is a specific subset of unknown or uncertain attributes to be estimated, the attribute vector can be partitioned as $z = (x, y)$, where $y \subseteq z$ denotes the target or output attributes. The target attributes can be different for different objects $z$. This scenario includes several usual inference paradigms. For instance, when there is a specific target symbolic attribute, the task is called *pattern recognition* or *classification*; when the target attribute is continuous, the inference task is called *regression* or *function approximation*. In general, we are interested in a general framework for *pattern completion* from partially known objects.

> Example 1: To illustrate this setting, assume that the preprocessor of a hypothetical computer vision system obtains features of a segmented object. The instances of the domain are described by the following $n=7$ attributes: AREA: $z^1 \in \mathbb{R}$, COLOR: $z^2 \in \{white, black, red, ...\}$, DISTANCE: $z^3 \in \mathbb{R}$, SHAPE: $z^4 \in \{circular, rectangular, triangular, ...\}$, TEXTURE: $z^5 \in \{soft, rough, ...\}$, OBJECTTYPE: $z^6 \in \{door, window, ...\}$ and ANGLE: $z^7 \in \mathbb{R}$. A typical instance may be $z = (78, blue, 3.4, triangular, soft, window, 45)$. If the object is partially occluded or 3-dimensional, some attributes will be missing or uncertain. For instance, the available information $S$ about $z$ could be expressed as (74±8, *blue* OR *black*, 3.4, *triangular*, ?, *window* 70% *door* 30%, ?), where $z^1$, $z^2$, $z^6$ are uncertain, $z^3$, $z^4$ are exact and $z^5$, $z^7$ are missing. In this case we could be interested in estimates for $y = \{z^5, z^6, z^7\}$ and even in improving our knowledge on $z^1$ and $z^2$.

The non-deterministic nature of many real world domains suggests a probabilistic approach, where attributes are considered as random variables. Objects are assumed to be drawn independently and identically distributed from $p(z) = p(z^1, ..., z^n) = p(x, y)$, the multivariate joint probability density function of the attributes, which completely characterizes the *n*-dimensional random variable $z$. To simplify notation, we will use the same function symbol $p(\cdot)$ to denote different p.d.f.'s if they can be identified without risk of confusion.

According to Statistical Decision Theory (Berger 1985), optimum estimators for the desired attributes are obtained through minimization of a suitable expected loss function:

$$\hat{y}_{OPT}(S) = argmin_{\hat{y}} \; \mathrm{E}\{L(y, \hat{y}) | S\}$$

where $L(y, \hat{y})$ is the loss incurred when the true $y$ is estimated by $\hat{y}$. Estimators are always features of the conditional or posterior distribution $p(y|S)$ of the target variables given the available information. For instance, the minimum squared error (MSE) estimator is the posterior *mean*, the minimum linear loss estimator is the posterior *median* and the minimum error probability (EP, 0-1 loss) estimator is the posterior *mode*.

> Example 2: A typical problem is the prediction of an unknown attribute $y$ from the observed attributes $x$. In this case the available information can be written as $S = (x, ?)$. If y is continuous, it is reasonable to use the MSE estimator: $\hat{y}_{MSE}(S) = E\{y \mid x\}$, the general regression function. If $y$ is symbolic and the same loss is associated to all errors, the EP estimator is adequate: $\hat{y}_{EP}(S) = argmax_y \, p(y|x) = argmax_y \, p(x|y)p(y)$. It corresponds to the *Maximum A Posteriori* rule or Bayes Test, widely used in Statistical Pattern Recognition.

The joint density $p(z) = p(x, y)$ plays an essential role in the inference process. It implicitly includes complete information about attribute dependences. In principle, any desired conditional distribution or estimator can be computed from the joint density by adequate integration. *Probabilistic Inference* is the process of computing the desired conditional probabilities from a (possibly implicit) joint distribution. From $p(z)$ (the prior, model of the domain, comprising implications) and $S$ (a known event, somewhat related to a certain $z$), we could obtain the posterior $p(z|S)$ and the desired target marginal $p(y|S)$ (the probabilistic "consequent").

> Example 3: If we observe an exact value $x_o$ in attribute $x$, i.e. $S = \{ x = x_o \}$, we have:

$$p(y|S) \equiv p(y \mid x_o) = \frac{p(x_o, y)}{\displaystyle\int_Y p(x_o, y) dy}$$

If we know that instance $z$ is in a certain region $R$ in the attribute space, i.e. $S = \{z \in R\}$, we compute the marginal density of $y$ from the joint $p(z) = p(x, y)$ restricted to region $R$ (Fig. 1):

$$p(\boldsymbol{y}|\boldsymbol{S}) = \int_X p(\boldsymbol{x}, \boldsymbol{y}|\{(\boldsymbol{x}, \boldsymbol{y}) \in R\})\, d\boldsymbol{x} = \frac{\int_R p(\boldsymbol{x}, \boldsymbol{y})\, d\boldsymbol{x}}{\iint_R p(\boldsymbol{x}, \boldsymbol{y})\, d\boldsymbol{x} d\boldsymbol{y}}$$

More general types of uncertain information $\boldsymbol{S}$ about $z$ will be discussed later.

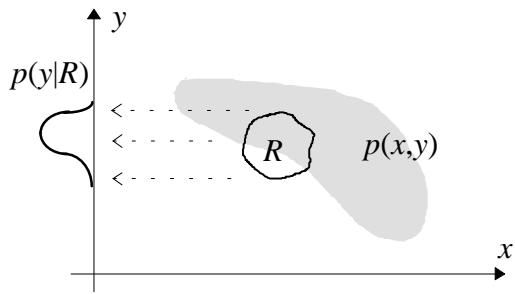

Figure 1. The conditional probability density of $\boldsymbol{y}$, assuming $z = (\boldsymbol{x}, \boldsymbol{y}) \in R$.

In summary, from the joint density $p(\boldsymbol{z})$ of a multivariate random variable, any subset of variables $\boldsymbol{y} \subseteq z$ may be, in principle, estimated given the available information $\boldsymbol{S}$ about the whole $z = (\boldsymbol{x}, \boldsymbol{y})$. In practical situations, two steps are required to solve the inference problem. First, a good model of the true joint density $p(\boldsymbol{z})$ must be obtained. Second, the available information $\boldsymbol{S}$ must be efficiently processed to improve our knowledge about future, partially specified instances $z$. These two complementary aspects, learning and inference, are approached from many scientific fields, providing different methodologies to solve practical applications.

From the point of view of Computer Science, the essential goal of Inductive Inference is to find an approximate intensional definition (properties) of an unknown concept (subset of the domain) from an incomplete extensional definition (finite sample). Machine Learning techniques (Michalski, Carbonell & Mitchell 1977, 1983, Hutchinson 1994) provide practical solutions (e.g. automatic construction of decision trees) to solve many situations where explicit programming must be avoided. Computational Learning Theory (Valiant 1993, Wolpert 1994, Vapnik 1995) studies the feasibility of induction in terms of generalization ability and resource requirements of different learning paradigms.

Under the general setting of Statistical Decision Theory, modeling techniques and the operational aspects of inference (based in numerical integration, Monte Carlo simulation, analytic approximations, etc.) are extensively studied from the Bayesian perspective (Berger 1985, Bernardo & Smith 1994). In the more specific field of Statistical Pattern Recognition (Duda & Hart 1973, Fukunaga 1990), standard parametric or nonparametric density approximation techniques (Izenman 1991) are used to learn from training data the class-conditional p.d.f.'s required by the optimum decision rule. For instance, if the class-conditional densities $p(\boldsymbol{x}|y)$ are Gaussian, the required parameters are the mean vector and covariance matrix of the feature vector in each class and the decision regions for $y$ in $\boldsymbol{x}$ have quadratic boundaries. Among the nonparametric classification techniques, the Parzen method and the K-N Nearest Neighbors rule must be mentioned. Analogously, if the target attribute is continuous and the statistical dependence between input and output variables $p(\boldsymbol{x},y)$ can be properly modeled by joint normality, we get multivariate linear regression: $\hat{y}_{MSE}(\boldsymbol{x}) = A\,\boldsymbol{x} + B$, where the required parameters are the mean values and the covariance matrix of the attrib-utes. Nonlinear regression curves can be also derived from nonparametric approximation techniques. Nonparametric methods present slower convergence rates, requiring significantly larger sample sizes to obtain satisfactory approximations; they are also strongly affected by the dimensionality of the data and the selection of the smoothing parameter is a crucial step. In contrast, they only require some kind of smoothness assumption on the target density.

Neural Networks (Hertz et. al 1991) are computational models trainable from empirical data that have been proposed to solve more complex situations. Their intrinsic parallel architecture is especially efficient in the inference stage. One of the most widely used neural models is the Multilayer Perceptron, a universal function approximator (Hornik et al. 1989) that breaks the limitations of linear decision functions. The Backpropagation learning algorithm (Rumelhart et al. 1986) can, in principle, adjust the network weights to implement arbitrary mappings, and the network outputs show desirable probabilistic properties (Wan 1990, Rojas 1996). There are also unsupervised networks for probability density function approximation (Kohonen 1989). However, neural models usually contain a large number of adjustable parameters, which is not convenient for generalization and, frequently, long times are required for training in relatively easy tasks. The input / output role of attributes cannot be changed in runtime and missing and uncertain values are poorly supported.

Bayesian Networks, based in the concept of conditional independence, are among the most relevant probabilistic inference technologies (Pearl 1988, Heckerman & Wellman 1995). The joint density of the variables is modeled by a directed graph which explicitly represents dependence statements. A wide range of inferences can be performed under this framework (Chang & Fung 1995, Lauritzen & Spiegelhalter 1988) and there are significant results on inductive learning of network structures (Bouckaert 1994, Cooper & Herskovits 1992, Valiveti & Oomen 1992). This approach is adequate when there is a large number of variables showing explicit dependences and simple cause-effect relations. Nevertheless, solving arbitrary queries is NP-Complete, automatic learning algorithms are time consuming and the allowed dependences between variables are relatively simple.

In an attempt to mitigate some of the above drawbacks, we have developed a general and efficient inference and learning framework based on the following considerations. It is well known (Titterington et al. 1985, McLachlan & Basford 1988, Dalal & Hall 1983, Bernardo & Smith 1994, Xu & Jordan 1996) that any reasonable probability density function $p(z)$ can be approximated up to the desired degree of accuracy by a finite mixture of simple components $C_i$, $i = 1..l$:

$$p(z) \cong \sum_i \mathrm{P}\{C_i\} \; p(z|C_i) \qquad (1)$$

The superposition of simple densities is extensively used to approximate arbitrary data dependences (Fig. 2). Maximum Likelihood estimators of the mixture parameters can be efficiently obtained from samples by the Expectation Maximization (EM) algorithm (Dempster, Laird & Rubin 1977, Redner & Walker 1984) (see Section 4).

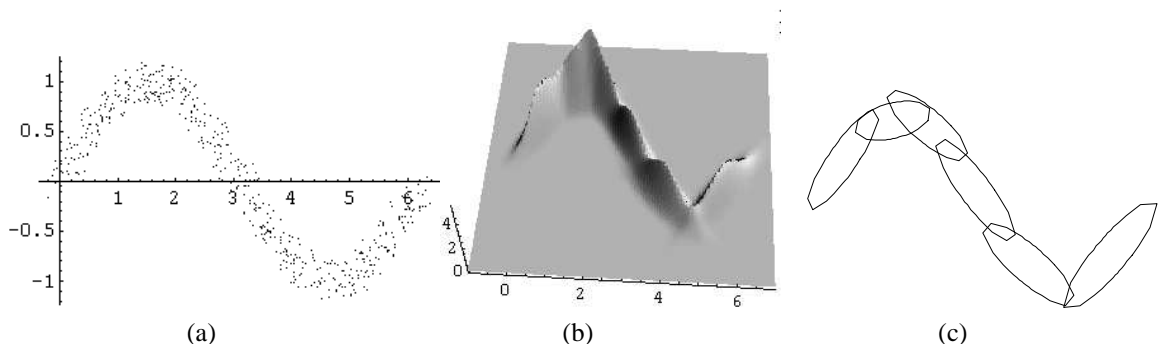

Figure 2. Illustrative example of density approximation using a mixture model. (a) Samples from a p.d.f. $p(x,y)$ showing a nonlinear dependence. (b) Mixture model for $p(x,y)$ with 6 gaussian components obtained by the standard EM algorithm. (c) Location of components.

The decomposition of probability distributions using mixtures has been frequently applied to unsupervised learning tasks, especially Cluster Analysis (McLachlan & Basford 1988, Duda & Hart 1973, Fukunaga 1990): the *a posteriori* probabilities of each postulated category are computed for all the examples, which are labeled according to the most probable source density. However, mixture models are specially useful in nonparametric supervised learning situations. For instance, the class conditional densities required in Statistical Pattern Recognition were individually approximated in (Priebe & Marchette 1991, Traven 1991) by finite mixtures; hierarchical mixtures of linear models were proposed in (Jordan & Jacobs 1994, Peng et. al 1995); mixtures of factor analyzers have been developed in (Ghahramani & Hinton 1996, Hinton, Dayan, & Revow 1997) and mixture models have been also useful for feature selection (Pudil et al. 1995). Mixture modeling is a growing semiparametric probabilistic learning methodology with applications in many research areas (Weiss & Adelson 1995, Fan et al. 1996, Moghaddam & Pentland 1997).

This paper introduces a framework for probabilistic inference and learning from arbitrary uncertain data: any piece of exact or uncertain information about both input and output values is consistently handled in the inference and learning stages. We approximate both the joint density $p(z)$ (model of the domain) and the relative likelihood function $p(S|z)$ (describing the available information) by a specific mixture model with factorized conjugate components, in such a way that numerical integration is avoided in the computation of any desired estimator, marginal or conditional density.

The advantages of modeling arbitrary densities using mixtures of natural conjugate components were already shown in (Dalal & Hall 1983), and, recently, inference procedures based in a similar idea have been proposed in (Ghahramani & Jordan 1994, Cohn et al. 1996, Peng et al. 1995, Palm 1994). However, our method efficiently handles uncertain data using explicit likelihood functions, which has not been extensively used before in Machine Learning, Pattern Recognition or related areas. We will follow standard probabilistic principles, providing natural statistical validation procedures.

The organization of the paper is as follows. Section 2 reviews some elementary results and concepts used in the proposed framework. Section 3 addresses the inference stage. Section 4 is concerned with learning, extending the EM algorithm to manage uncertain information. Section 5 discusses the method in relation to alternative techniques and presents experimental evaluation. The last section summarizes the conclusions and future directions of this work.

## 2. Preliminaries

### 2.1 A Calculus of Generalized Normals

In many applications, the instances of the domain are represented simultaneously by continuous and symbolic or discrete variables (as in Wilson & Martinez 1997). To simplify notation, we will denote both probability impulses and Gaussian densities by means of a common formalism. The *generalized normal* $\mathcal{N}(x,\mu,\sigma)$ denotes a probability density function with the following properties:

If $\sigma > 0$,
$$\mathcal{N}(x,\mu,\sigma) \equiv \frac{1}{\sqrt{2\pi}\sigma} exp\left[\frac{-(x-\mu)^2}{2\sigma^2}\right]$$

If $\sigma = 0$,
$$\mathcal{N}(x,\mu,\sigma) = \mathcal{N}(x,\mu,0) \equiv \mathcal{N}(x,\mu) \equiv \delta(x-\mu)$$

$\mathcal{N}(x,\mu,\sigma)$ is a Gaussian density with mean $\mu$ and standard deviation $\sigma \geq 0$. When the dispersion is zero, $\mathcal{N}$ reduces to a Dirac's delta function located at $\mu$. In both cases $\mathcal{N}$ is a proper p.d.f.:

$$\int_X \mathcal{N}(x,\mu,\sigma)\, dx = 1 \qquad \mathcal{N}(x,\mu,\sigma) > 0$$

The product of generalized normals can be elegantly expressed (Papoulis 1991 pp. 258, Berger 1985) by:

for $\sigma_1 + \sigma_2 > 0$:
$$\mathcal{N}(x,\mu_1,\sigma_1) \cdot \mathcal{N}(x,\mu_2,\sigma_2) = \mathcal{N}(x,\eta,\varepsilon) \cdot \mathcal{N}(\mu_1,\mu_2, \sqrt{\sigma_1^2 + \sigma_2^2}) \qquad (2)$$

where the mean $\eta$ and dispersion $\varepsilon$ of the new normal are given by:

$$\eta = \frac{\sigma_1^2 \mu_2 + \sigma_2^2 \mu_1}{\sigma_1^2 + \sigma_2^2} \qquad \varepsilon^2 = \frac{\sigma_1^2 \sigma_2^2}{\sigma_1^2 + \sigma_2^2}$$

This relation is useful for computing the integral of the product of two generalized normals:

for $\sigma_1 + \sigma_2 > 0$:
$$\int_X \mathcal{N}(x,\mu_1,\sigma_1) \cdot \mathcal{N}(x,\mu_2,\sigma_2)\, dx = \mathcal{N}(\mu_1, \mu_2, \sqrt{\sigma_1^2 + \sigma_2^2}) \qquad (3)$$

And, for consistency, we define

for $\sigma_1 = \sigma_2 = 0$:
$$\int_X \mathcal{N}(x,\mu_1)\, \mathcal{N}(x,\mu_2)\, dx = \mathcal{N}(\mu_1,\mu_2) \equiv I\{\mu_1 = \mu_2\}$$

where $I\{predicate\} = 1$ if *predicate* is true and zero otherwise. Virtually any reasonable univariate probability distribution or likelihood function can be accurately modeled by an appropriate mixture of generalized normals. In particular, p.d.f.'s over symbolic variables are mixtures of impulses. Without loss of generality, symbols may be arbitrarily mapped to specific numbers and represented over numeric axes. Integrals over discrete domains become sums.

> Example 4: Let us approximate the p.d.f. $p(x)$ of a mixed continuous and symbolic valued random variable $x$ by a mixture of generalized normals. Assume that $x$ takes with probability 0.4 the exact value 10 (with a special meaning), and with probability 0.6 a random value continuously distributed following the triangular shape shown in Fig. 3. The density $p(x)$ can be accurately approximated (see Section 4) using 4 generalized normals:
>
> $$p(x) \cong .40\mathcal{N}(x,10) + .21\mathcal{N}(x,.04,.23) + .28\mathcal{N}(x,.45,.28) + .11\mathcal{N}(x,.99,.21)$$

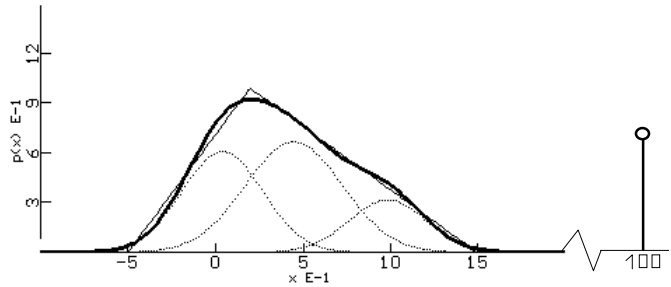

Figure 3. The p.d.f. of a mixed random variable approximated by a mixture of generalized normals.

## 2.2 Modeling Uncertainty: The Likelihood Principle

Assume that the value of a random variable $z$ must be inferred from a certain observation or subjective information $S$. If $z$ has been drawn from $p(z)$ and the measurement or judgment process is characterized by a conditional $p(S|z)$, our knowledge on $z$ is updated according to $p(z|S)=p(z)\,p(S|z)\,/\,p(S)$, where $p(S) = \int_Z p(S|z)\,p(z)\,dz$  (see Fig. 4).

The likelihood function $f_S(z) \equiv p(S|z)$ is the probability density ascribed to $S$ by each possible $z$. It is an arbitrary nonnegative function over $z$ that can be interpreted in two alternative ways. It can be the "objective" conditional distribution $p(S|z)$ of a physical measurement process (e.g. a model of sensor noise, specifying bias and variance of the observable $S$ for every possible true value $z$), also known as *error model*. It can also be a "subjective" judgment about the chance of the different $z$ values (e.g. intervals, more likely regions, etc.), based on vague or difficult to formalize information. The dispersion of $f_S(z)$ is directly related to the uncertainty associated to the measurement process. Following the *likelihood principle* (Berger 1985), we explicitly assume that all the experimental information required to perform probabilistic inference is contained in the likelihood function $f_S(z)$.

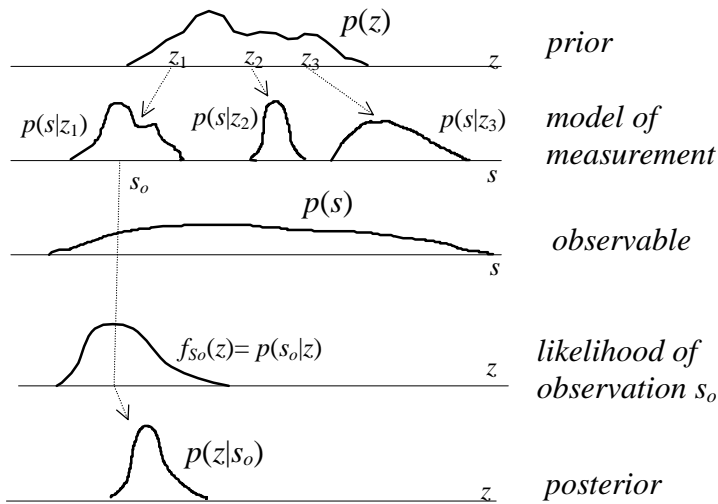

Figure 4. Illustration of the elementary Bayesian univariate inference process.

## 2.3 Inference Using Mixtures of Conjugate Densities

The computation of $p(z|S)$ may be hard, unless $p(z)$ and $p(S|z)$ belong to special (conjugate) families (Berger 1985, Bernardo & Smith 1994). In this case the posterior density can be analytically obtained from the parameters of the prior and the likelihood, avoiding numeric integration. The prior, the likelihood and the posterior are in the same mathematical family. The "belief structure" is closed under the inference process.

> Example 5: In the univariate case, assume that $z$ is known to be normally distributed around $r$ with dispersion $\sigma_r$, i.e. $p(z) = \mathcal{N}(z, r, \sigma_r)$. Assume also that our measurement device has Gaussian noise, so the observed values are distributed according to $p(s|z) = \mathcal{N}(s, z, \sigma_s)$. Therefore, if we observe a certain value $s_o$, from the property of the product of generalized normals in eq. (2), the posterior knowledge on $z$ becomes another normal $\mathcal{N}(z, \mu, \varepsilon)$. The new expected location of $z$ can be expressed as a weighted average of $r$ and $s_o$: $\eta = \gamma \, s_o + (1-\gamma) r$ and the uncertainty is reduced to $\varepsilon^2 = \gamma \, \sigma_s^2$. The coefficient $\gamma = \sigma_r^2 / (\sigma_r^2 + \sigma_s^2)$ quantifies the relative importance of the experiment with respect to the prior.

This computational advantage can be extended to the general case by using mixtures of conjugate families (Dalal & Hall 1983) to approximate the desired joint probability distribution and the likelihood function.

> Example 6: If the domain and the likelihood are modeled respectively by
>
> $$p(z) \cong \sum_i P_i \, \mathcal{N}(z, \mu_i, \sigma_i) \qquad p(s_o|z) \cong \sum_r \pi_r \, \mathcal{N}(z, \eta_r, \varepsilon_r)$$
>
> (where $\pi_r$, $\eta_r$ and $\varepsilon_r$ depend explicitly on the observed $s_o$), then the posterior can be also written as the following mixture:
>
> $$p(z|s_o) \cong \sum_{i,r} \theta_{i,r} \, \mathcal{N}(z, \nu_{i,r}, \lambda_{i,r}) \qquad (4)$$
>
> From properties (2) and (3), the parameters $\nu_{i,r}$ and $\lambda_{i,r}$ and the weights $\theta_{i,r}$ are given by:
>
> $$\nu_{i,r} = \frac{\sigma_i^2 \eta_r + \varepsilon_r^2 \mu_i}{\sigma_i^2 + \varepsilon_r^2} \qquad \lambda_{i,r} = \frac{\sigma_i \, \varepsilon_r}{\sqrt{\sigma_i^2 + \varepsilon_r^2}}$$
>
> $$\theta_{i,r} \equiv \frac{P_i \pi_r \, \mathcal{N}(\mu_i, \eta_r, \sqrt{\sigma_i^2 + \varepsilon_r^2})}{\sum_{k,l} P_k \pi_l \, \mathcal{N}(\mu_k, \eta_l, \sqrt{\sigma_k^2 + \varepsilon_l^2})}$$

## 2.4 The Role of Factorization

Given a multivariate observation $z$ partitioned into two subvectors, $z = (x, y)$, assume that we are interested in inferring the value of the unknown attributes $y$ from the observed attributes $x$. Note that if $x$ and $y$ are statistically independent, the joint density is factorizable: $p(z) = p(x, y) = p(x) p(y)$ and, therefore, the posterior $p(y|x)$ equals the prior marginal $p(y)$. The observed $x$ carries no predictive information about $y$ and the optimum estimators do no depend on $x$. For instance, $\hat{y}_{MSE}(x) = \mathrm{E}\{y|x\} = \mathrm{E}\{y\}$ and $\hat{y}_{EP}(x) = argmax_y \, p(y)$. This is the simplest estimation task. No runtime computations are required for the optimum solution, which may be precalculated.

In realistic situations the variables are statistically dependent. In general, the joint density cannot be factorized and the required marginal densities may be hard to compute. However, interesting consequences arise if the joint density is expressed as a finite mixture of factorized (with independent variables) components $C_1, C_2, ..., C_l$:

$$p(\mathbf{z}) = p(z^1, ..., z^n) = \sum_i \mathrm{P}\{C_i\}\ p(\mathbf{z}|C_i) = \sum_i \mathrm{P}\{C_i\} \prod_j p(z^j|C_i) \qquad (5)$$

This structure is convenient for inference purposes. In particular, in terms of the desired partition of $\mathbf{z} = (\mathbf{x}, \mathbf{y})$:

$$p(\mathbf{z}) = p(\mathbf{x}, \mathbf{y}) = \sum_i \mathrm{P}\{C_i\}\ p(\mathbf{x}|C_i)\ p(\mathbf{y}|C_i)$$

so the marginal densities are mixtures of the marginal components:

$$p(\mathbf{x}) = \int_Y p(\mathbf{x}, \mathbf{y}) dy = \sum_i \mathrm{P}\{C_i\}\ p(\mathbf{x}|C_i)$$

$$p(\mathbf{y}) = \sum_i \mathrm{P}\{C_i\}\ p(\mathbf{y}|C_i)$$

and the desired conditional densities are also mixtures of the marginal components:

$$p(\mathbf{y}|\mathbf{x}) = \sum_i \alpha_i(\mathbf{x})\ p(\mathbf{y}|C_i) \qquad (6)$$

where the weights $\alpha_i(\mathbf{x})$ are the probabilities that the observed $\mathbf{x}$ has been generated by each component $C_i$:

$$\alpha_i(\mathbf{x}) = \frac{\mathrm{P}\{C_i\}\,p(\mathbf{x}|C_i)}{\sum_j \mathrm{P}\{C_j\}\,p(\mathbf{x}|C_j)} = \mathrm{P}\{C_i|\mathbf{x}\}$$

The p.d.f. approximation capabilities of mixture models with factorized components remain unchanged, at the cost of a possibly higher number of components to obtain the desired degree of accuracy, avoiding "artifacts" (see Fig. 5). Section 5.2 discusses the implications of factorization in relation with alternative model structures.

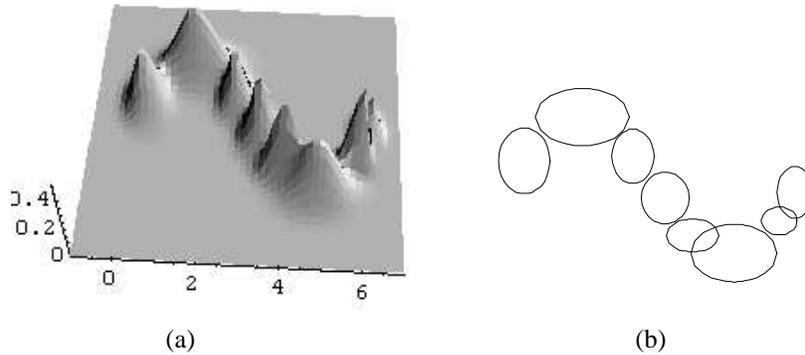

(a)                                                          (b)

Figure 5. (a) Density approximation for the data in Fig. 2, using a mixture with 8 factorized components. (b) Location of components. Note how an arbitrary dependence can be represented as a mixture of components which itself have independent variables (observe that a somewhat "smoother" solution could be obtained increasing the number of components).

## 3. The MFGN Framework

The previous concepts will be integrated in a general probabilistic inference framework that we call MFGN (*Mixtures of Factorized Generalized Normals*). Fig. 6 shows the abstract dependence relations among attributes in a generic domain (upper section of the figure) and between the attributes and the observed information (lower section). In the MFGN framework, both relations are modeled by finite mixtures of products of generalized normals. The key idea is using factorization to cope with multivariate domains and heterogeneous attribute vectors, and conjugate densities to efficiently perform inferences given arbitrary uncertain information. In this section, we will derive the main inference expressions. The learning stage will be described in Section 4.

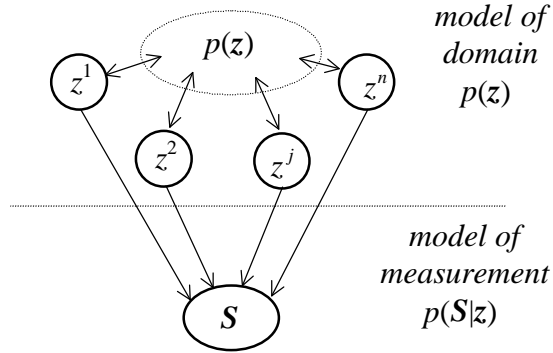

Figure 6. Generic dependences in the inference process.

### 3.1 Modeling Attribute Dependences in the Domain

In the MFGN framework the attribute dependencies in the domain are modeled by a joint density in the form of a finite mixture of factored components, as in expression (5), where the component marginals $p(z^j|C_i) \equiv \mathcal{N}(z^j,\mu_i^j,\sigma_i^j)$ are generalized normals:

$$p(z) = \sum_i P_i \prod_j \mathcal{N}(z^j,\mu_i^j,\sigma_i^j) \qquad i=1..l, \; j=1..n, \qquad (7)$$

If desired, the terms associated to the pure symbolic attributes $z^j$ (with all the $\sigma_i^j = 0$) can be collected in such a way that the component marginals are expressed as mixtures of impulses:

$$p(z^j|C_i) \equiv \sum_\omega t_{i,\omega}^j \mathcal{N}(z^j,\omega) \qquad (8)$$

where $t_{i,\omega}^j \equiv \mathrm{P}\{z^j = \omega|C_i\}$ is the probability that $z^j$ takes its $\omega$-th value in component $C_i$. This manipulation reduces the number $l$ of global components in the mixture. The adjustable parameters of the model are the proportions $P_i = \mathrm{P}\{C_i\}$ and the mean value $\mu_i^j$ and dispersion $\sigma_i^j$ of the $j$-th attribute in the $i$-th component (or, for the symbolic attributes, the probabilities $t_{i,\omega}^j$). While the structure (8) will be explicitly used for symbolic attributes in applications and illustrative examples, most of the mathematical derivations will be made over the concise expression (7).

When all variables are continuous, the MFGN architecture reduces to a mixture of gaussians with diagonal covariance matrices. The proposed factorized structure extends the properties of diagonal covariance matrices to heterogeneous attribute vectors. We are interested in joint models, which support inferences from partial information about any subset of variables. Note that there is not an easy way to define a measure of statistical depencence between symbolic and continuous attributes, to be used as a parameter of some probability density function[1]. The required "heterogeneous" dependence model can be conveniently captured by superposition of simple factorized (with independent variables) densities.

Example 7: Figure 7 shows an illustrative 3-attribute data set ($x$ and $y$ are continuous and $z$ is symbolic) and the components of the MFGN approximation obtained by the EM algorithm (see Section 4) for their joint density. The parameters of the mixture are shown in Table 1. Note that, because of the overlapped structure of the data, some components (5 and 6) are "assigned" to both values of the symbolic attribute $z$.

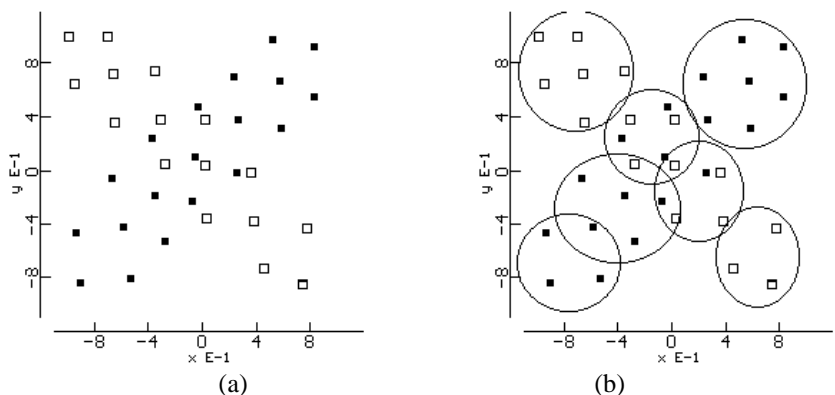

(a)        (b)

Figure 7. (a) Simple data set with two continuous and one symbolic attribute.
(b) Location of the mixture components.

| $i$ | $P_i$ | $\mu_i^x$ | $\sigma_i^x$ | $\mu_i^y$ | $\sigma_i^y$ | $t_{i.white}^z$ | $t_{i.black}^z$ |
|---|---|---|---|---|---|---|---|
| 1 | .14 | −.40 | .24 | −.27 | .20 | 0 | 1 |
| 2 | .09 | −.76 | .19 | −.68 | .18 | 0 | 1 |
| 3 | .20 | .55 | .23 | .66 | .24 | 0 | 1 |
| 4 | .17 | −.71 | .27 | .76 | .22 | 1 | 0 |
| 5 | .13 | .21 | .17 | −.14 | .19 | .74 | .26 |
| 6 | .18 | −.14 | .18 | .26 | .17 | .55 | .45 |
| 7 | .09 | .65 | .16 | −.64 | .19 | 1 | 0 |

Table 1. Parameters of the Mixture Model for the Data Set in Fig. 7.

## 3.2 Modeling Arbitrary Information about Instances

The available information about a particular instance $z$ is denoted by $S$. Following the likelihood principle, we are not concerned with the true nature of $S$, whether it is some kind of physical meas-urement or a more subjective judgment about the location of $z$ in the attribute space. All we need to update our knowledge about $z$, in the form of the posterior $p(z|S)$, is the relative likelihood function $p(S|z)$ of the observed $S$. In general, $p(S|z)$ can be any nonnegative multivariable function $f_S(z)$ over the domain. In the objective case, statistical studies of the measurement process can be used to determine the likelihood function. In the subjective case, it may be obtained by standard distribution elicitation techniques (Berger 1985). In either case, under the MFGN framework, the likelihood function of the available information to be used in the inference process will be approximated, up to the desired degree of accuracy, by a sum of products of generalized normals:

$$p(S/z) = \sum_r P\{S|s_r\} p(s_r|z) = \sum_r P\{S|s_r\} \prod_j p(s_r^j|z^j)$$

$$= \sum_r \pi_r \prod_j \mathcal{N}(z^j, s_r^j, \varepsilon_r^j) \qquad (9)$$

---

[1] For this reason, in pattern classification tasks separate models are typically built for each class-conditional density.

Without loss of generality, the available knowledge is structured as a weighted disjunction $S = \{\pi_1 s_1 \text{ OR } \pi_2 s_2 \dots \text{ OR } \pi_R s_R\}$ of conjunctions $s_r = \{s_r^1 \text{ AND } s_r^2 \dots \text{ AND } s_r^n\}$ of elementary uncertain observations in the form of generalized normal likelihoods $\mathcal{N}(z^j, s_r^j, \varepsilon_r^j)$ centered at $s_r^j$ with uncertainty $\varepsilon_r^j$. The measurement process can be interpreted as the result of $R$ (objective or subjective) sensors $s_r$, providing conditionally independent information $p(s_r^j|z^j)$ about the attributes (each $s_r^j$ only depends on $z^j$) with relative strength $\pi_r$. Note that any complex uncertain information about an instance $z$, expressed as a nested combination of elementary uncertain beliefs $s_r^j$ about $z^j$ using "probabilistic connectives", can be ultimately expressed by structure (9) (OR translates to addition, AND translates to product and the product of two generalized normals over the same attribute becomes a single, weighted normal).

Example 8: Consider the hypothetical computer vision domain in Example 1. Assume that the information about an object $z$ is the following: "AREA is around $a$ and DISTANCE is around $b$ or, more likely, SHAPE is surely *triangular* or else *circular* and AREA is around $c$ and ANGLE is around $d$ or equal to $e$". This structured piece of information can be formalized as:

$$p(S|z) = .3 \, [\mathcal{N}(z^1, a, \varepsilon_a) \, \mathcal{N}(z^3, b, \varepsilon_b)]$$

$$+ .7 \, [ \, (.9\mathcal{N}(z^4, triang) + .1\mathcal{N}(z^4, circ)) \, \mathcal{N}(z^1, c, \varepsilon_c) \, (\mathcal{N}(z^7, d, \varepsilon_d) + \mathcal{N}(z^7, e)) \, ]$$

which, expanded, becomes the mixture of 5 factorized components operationally represented by the parameters shown in Table 2.

In a simpler situation, the available information about z could be a conjunction of uncertain attributes similar to {Color = red 0.8 green 0.2} and {Area = 3 ± .5} and {Shape = rectangular 0.6 circular 0.3 triangular 0.1}. The likelihood of Shape values can be obtained from the output of a simple pattern classifier (e.g. K-N-nearest neighbors) over moment invariants, while attributes as Color and Area are directly extracted from the image. In this case we could be interested in the distribution of values for other attributes as Texture and ObjectType. Alternatively, we could start from {ObjectType = door 0.6 window 0.4} and {Texture = rough} in order to determine the probabilities of Color and Angle values for selecting a promising search region.

| $\pi_r$ | $s_r^1,\varepsilon_r^1$ | $s_r^2,\varepsilon_r^2$ | $s_r^3,\varepsilon_r^3$ | $s_r^4,\varepsilon_r^4$ | $s_r^5,\varepsilon_r^5$ | $s_r^6,\varepsilon_r^6$ | $s_r^7,\varepsilon_r^7$ |
|---|---|---|---|---|---|---|---|
| .30 | **$a,\varepsilon_a$** | -, ∞ | **$b,\varepsilon_b$** | -, ∞ | -, ∞ | -, ∞ | -, ∞ |
| .63 | **$c,\varepsilon_c$** | -, ∞ | -, ∞ | **triang, 0** | -, ∞ | -, ∞ | **$d,\varepsilon_d$** |
| .63 | **$c,\varepsilon_c$** | -, ∞ | -, ∞ | **triang, 0** | -, ∞ | -, ∞ | **e, 0** |
| .07 | **$c,\varepsilon_c$** | -, ∞ | -, ∞ | **circ, 0** | -, ∞ | -, ∞ | **$d,\varepsilon_d$** |
| .07 | **$c,\varepsilon_c$** | -, ∞ | -, ∞ | **circ, 0** | -, ∞ | -, ∞ | **e, 0** |

Table 2. Parameters of the Uncertain Information Model in Example 8.

## 3.3  The Joint Model-Observation Density

The generic dependence structure in Fig. 6 is implemented by the MFGN framework as shown in Fig. 8. The upper section of the figure is the model of nature, obtained in a previous learning stage and used for inference without further changes. Dependences among attributes are conducted through an intermediary hidden or latent component $C_i$. The lower section represents the available uncertain information, measurement model or query structure associated to each particular inference operation.

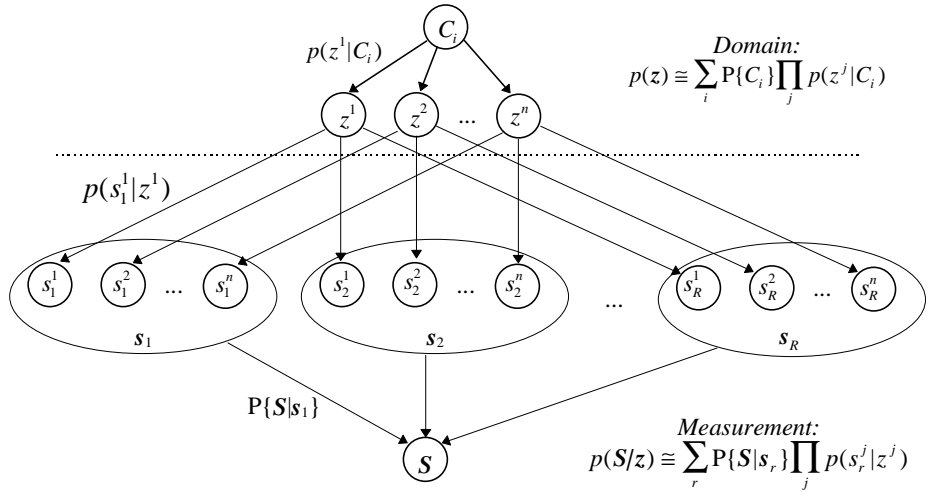

Figure 8. Structure of the MFGN model. The attributes are conditionally independent. The measurement process is modeled by a collection of independent "virtual" sensors $p(s_r^j|z^j)$.

The joint density of the relevant variables becomes:

$$p(C_i,z,s_r,S) = \mathrm{P}\{S|s_r\}\ p(s_r|z)\ p(z|C_i)\ \mathrm{P}\{C_i\}$$

$$= \mathrm{P}\{C_i\}\ \mathrm{P}\{S|s_r\}\prod_j p(s_r^j|z^j)\ p(z^j|C_i)$$

$$= P_i\ \pi_r\prod_j \mathcal{N}(z^j,s_r^j,\varepsilon_r^j)\ \mathcal{N}(z^j,\mu_i^j,\sigma_i^j) \tag{10}$$

Now we will derive an alternative expression for eq. (10) which is more convenient for computing the marginal densities of any desired variable. Using the following relation:

$$p(s_r^j|z^j) \ p(z^j|C_i) = p(z^j, s_r^j|C_i) = p(z^j|s_r^j, C_i) \ p(s_r^j|C_i)$$

and properties (2) and (3), we define the "dual" densities of the model:

$$\beta_{i,r}^j \equiv p(s_r^j|C_i) = \int_{Z^j} p(s_r^j|z^j) \ p(z^j|C_i) \ dz^j = \mathcal{N}(s_r^j, \mu_i^j, \rho_{i,r}^j) \tag{11}$$

$$\psi_{i,r}^j(z^j) \equiv p(z^j|s_r^j, C_i) = \frac{p(s_r^j|z^j)}{p(s_r^j|C_i)} \ p(z^j|C_i) = \mathcal{N}(z^j, \nu_{i,r}^j, \lambda_{i,r}^j) \tag{12}$$

where the parameters $\rho_{i,r}^j$, $\nu_{i,r}^j$ and $\lambda_{i,r}^j$ are given by:

$$\rho_{i,r}^j \equiv \sqrt{(\sigma_i^j)^2 + (\varepsilon_r^j)^2}$$

$$\nu_{i,r}^j \equiv \frac{(\sigma_i^j)^2 s_r^j + (\varepsilon_r^j)^2 \mu_i^j}{(\rho_{i,r}^j)^2}$$

$$\lambda_{i,r}^j \equiv \frac{\sigma_i^j \varepsilon_r^j}{\rho_{i,r}^j}$$

$\beta_{i,r}^j$ is the likelihood of the $r$-th elementary observation $s_r^j$ of the $j$-th attribute $z^j$ in each component $C_i$ and $\psi_{i,r}^j(z^j)$ is the effect of the $r$-th elementary information $s_r^j$ about the $j$-th attribute $z^j$ over the marginal component $p(z^j|C_i)$ in each component $C_i$. Using the above notation, the MFGN model structure can be conveniently written as:

$$p(C_i, z, s_r, S) = P_i \ \pi_r \prod_j \beta_{i,r}^j \ \psi_{i,r}^j(z^j) \tag{13}$$

### 3.4 The Posterior Density

In the inference process the available information is combined with the model of the domain to update our knowledge about a particular object. Given a new piece of information $S$ we must compute the posterior distribution $p(y|S)$ of the desired target attributes $y \subseteq z$. Then, estimators $\hat{y}(S) \cong y$ can be obtained from $p(y|S)$ to minimize any suitable average loss function. This is efficiently supported under the MFGN framework regardless of the complexity of the domain $p(z)$ and the structure of the available information $S = \{\pi_r, s_r\}$.

The attributes are partitioned into two subvectors $z = (x, y)$, where $y = \{z^d\}$ are the desired target attributes and $x = \{z^o\}$ are the rest of attributes. Accordingly, each component $s_r$ of the available information $S$ is partitioned as $s_r = (s_r^x, s_r^y)$. The information about the target attributes $y$ in the $r$-th observation, independent from the model $p(z)$, is denoted by $s_r^y$ (often $y$ is just missing and there are no such pieces of information) and $s_r^x$ represents the information about the rest of attributes $x$. Using this convention we can write:

$$p(z, S) = p(x, y, S) = \sum_{i,r} P_i \, \pi_r \, \beta_{i,r} \, \Psi_{i,r}(x) \, \Psi_{i,r}(y)$$

Where $\beta_{i,r}$ is the likelihood of the $r$-th conjunction $s_r$ of $S$ in component $C_i$ :

$$\beta_{i,r} \equiv \prod_j \beta_{i,r}^j \tag{14}$$

and the terms $\psi_{i,r}^j(z^j)$ are grouped according to the partition of $z = (x, y)$:

$$\Psi_{i,r}(x) \equiv \prod_o \psi_{i,r}^o(z^o) \qquad\qquad \Psi_{i,r}(y) \equiv \prod_d \psi_{i,r}^d(z^d)$$

The desired posterior $p(y|S) = p(y,S) \, / \, p(S)$ can be computed from the joint $p(z,S)$ by marginalization: along $x$ to obtain $p(y,S)$ and along all $z$ to obtain $p(S)$. Note that each univariate marginalization of $p(z,S)$ along attribute $z^j$ eliminates all the terms $\psi_{i,r}^j(z^j)$ in the sum (13):

$$p(y, S) = \int_X p(x, y, S) \, dx = \sum_{i,r} P_i \, \pi_r \, \beta_{i,r} \, \Psi_{i,r}(y)$$

$$p(S) = \int_Z p(z, S) \, dz = \sum_{i,r} P_i \, \pi_r \, \beta_{i,r}$$

Therefore, the posterior density can be compactly written as:

$$p(y|S) = \sum_{i,r} \alpha_{i,r} \Psi_{i,r}(y) \tag{15}$$

where $\alpha_{i,r}$ is the probability that the object $z$ has been generated by component $C_i$ and the elementary information $s_r$ is true, given the total information $S$:

$$\alpha_{i,r} \equiv \mathrm{P}\{C_i, s_r | S\} = \frac{P_i \, \pi_r \, \beta_{i,r}}{\sum_{k,l} P_k \, \pi_l \, \beta_{k,l}} \tag{16}$$

and $\Psi_{i,r}(y) = p(y|s_r^y, C_i)$ is the marginal density $p(y|C_i)$ of the desired attributes in the $i$-th component, modified by the contribution of all the associated $s_r^y$. Since $p(y|s_r^y, C_i) = p(y|s_r, C_i)$, the expression (16) also follows from the expansion:

$$p(y|S) = \sum_{i,r} p(y|s_r, C_i) \, \mathrm{P}\{C_i, s_r | S\}$$

In summary, when the joint density and the likelihood function are approximated by mixture models with the proposed structure, the computation of conditional densities given events of arbitrary "geometry" is notably simplified. Factorized components reduce multidimensional integration to simple combination of univariate integrals and conjugate families avoid numeric integration. This property is illustrated in Fig. 9.

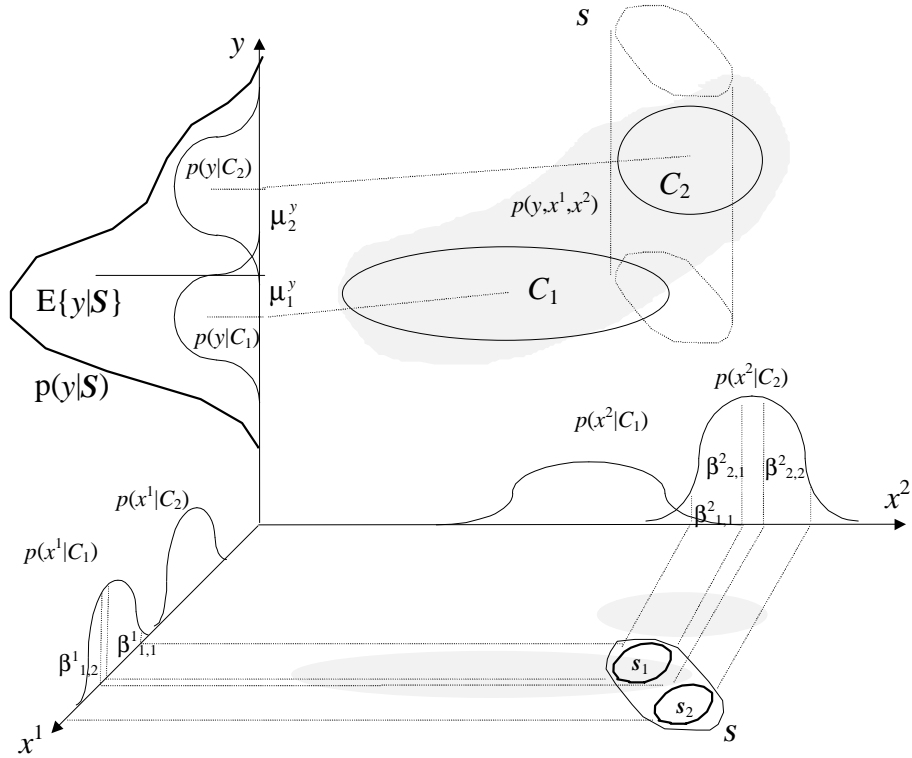

Figure 9. Graphical illustration of the essential property of the MFGN framework. Consider the MSE estimate for $y$, conditioned to the event that $(y, x^1, x^2)$ is in the cylindrical region $S$. The required multidimensional integrations are computed analytically in terms of the marginal likelihoods $\beta^j_{i,r}$ associated to each attribute and each pair of components $C_i$ and $s_r$ of the models for $p(y, x^1, x^2)$ and for $S$, respectively. In this case $\Psi_{i,r}(y) = p(y|C_i)$ because no information about $y$ is supplied in $S$.

<u>Example 9</u>: Fig. 10.a shows the joint density of two continuous variables $x$ and $y$. It is modeled as a mixture with 30 factorized generalized normals. Fig. 10.b shows the likelihood function of the event $S_1 = \{(x \equiv y \text{ OR } x \equiv -y) \text{ AND } y > 0\}$. Fig. 10.c shows the posterior joint density $p(x,y|S_1)$. Fig. 10.d shows the likelihood function of the event $S_2 = \{(x,y) \cong (0,0) \text{ OR } x \equiv 3\}$. Fig. 10.e shows the posterior joint density $p(x,y|S_2)$. Fig. 10.f and 10.g show respectively the posterior marginal density $p(x|S_2)$ and $p(y|S_2)$. These complex inferences are analytically computed under the MFGN framework, without any numeric integration.

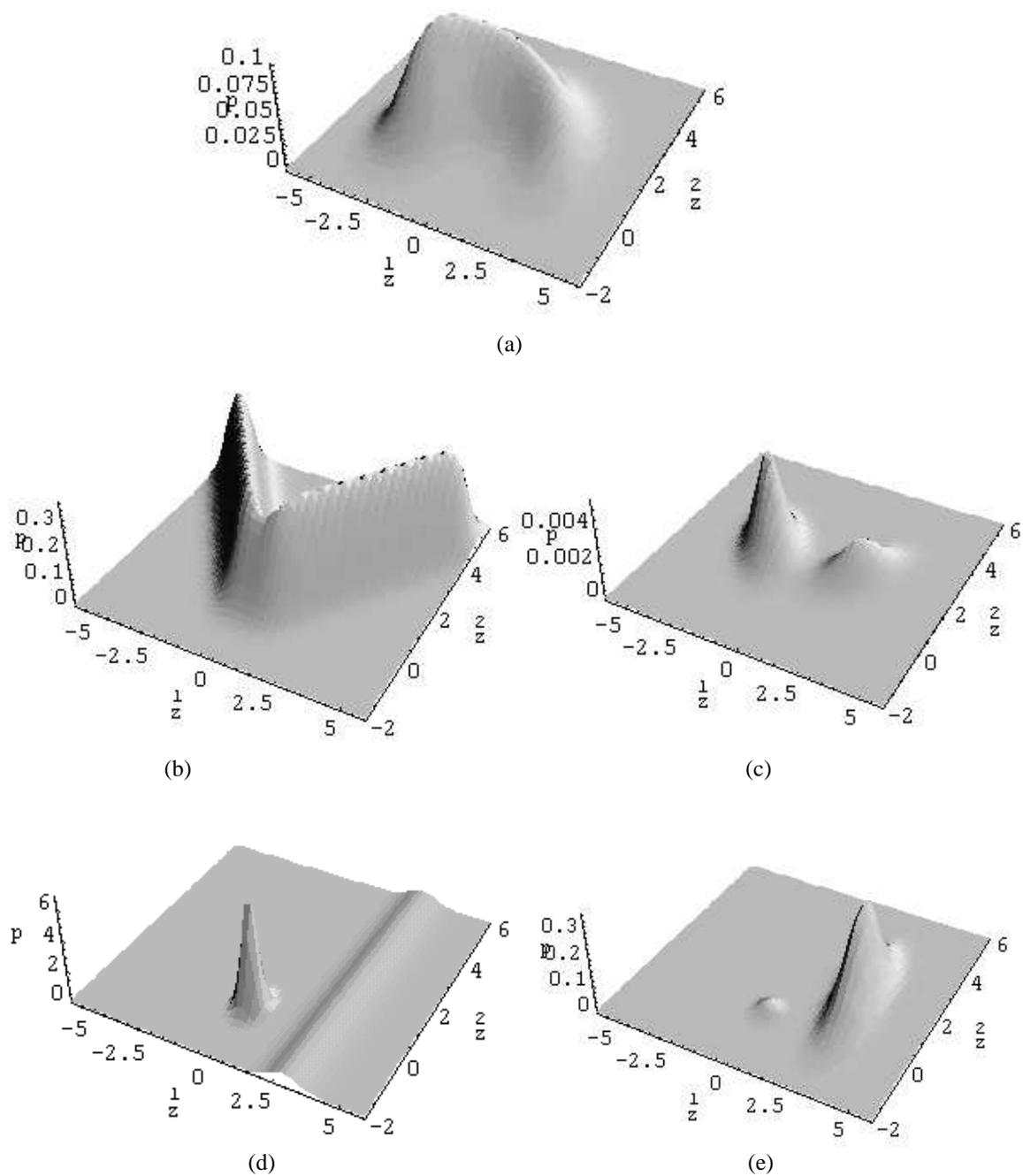

Figure 10. Illustrative examples of probabilistic inference from arbitrary uncertain information in the MFGN framework (see Example 9).

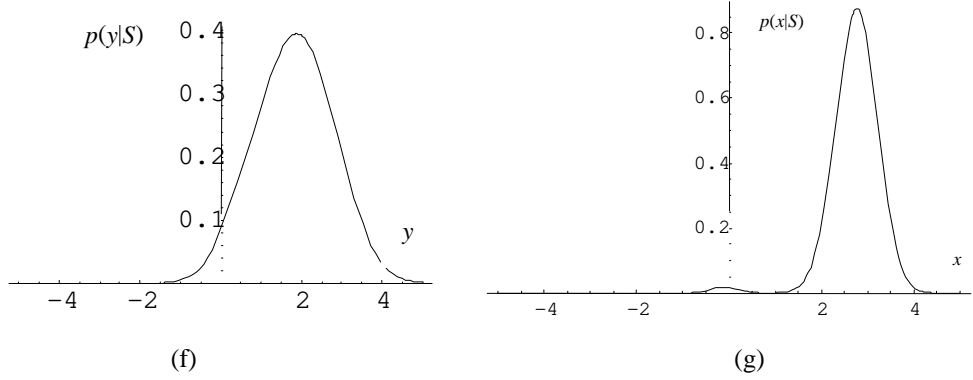

(f)                        (g)

Figure 10. (cont.).

### 3.5 Expressions for the Estimators

Approximations to the optimum estimators can be easily obtained by taking advantage of the mathematically convenient structure of the posterior density. Under the MFGN framework, the conditional expected value of any function $g(y)$ becomes a linear combination of constants:

$$E\{g(y)|S\} = \int_Y g(y)\ p(y|S)\ dy =$$

$$= \sum_{i,r} \alpha_{i,r} \int_Y g(y)\ \Psi_{i,r}(y)\ dy\ =\ \sum_{i,r} \alpha_{i,r}\ E_{i,r}\{g(y)\} \tag{17}$$

where $E_{i,r}\{g(y)\} \equiv E\{g(y)|s_r^y, C_i\}$ is the expected value of $g(y)$ in the $i$-th component[2] modified[3] by the $r$-th observation $s_r^y$ :

$$E_{i,r}\{g(y)\} \equiv \int_Y g(y) \prod_d \mathcal{N}(z^d, \nu_{i,r}^d, \lambda_{i,r}^d)\ dy$$

We can now analytically compute the desired optimum estimators. For instance, the MSE estimator for a single continuous attribute $y = z^d$ requires the mean values $E_{i,r}\{z^d\} = \nu_{i,r}^d$ :

$$\hat{y}_{MSE}(S) = E\{y|S\} = \sum_{i,r} \alpha_{i,r}\ \nu_{i,r}^d$$

From our explicit expression for $p(y|S)$ we can also compute the conditional cost:

$$e_{MSE}^2(S) = E\Big\{\big(y - \hat{y}_{MSE}(S)\big)^2|S\Big\} = E\{y^2|S\} - \hat{y}_{MSE}^2(S) =$$

$$= \sum_{i,r} \alpha_{i,r}\left[(\nu_{i,r}^d)^2 + (\lambda_{i,r}^d)^2\right] - \left(\sum_{i,r} \alpha_{i,r}\ \nu_{i,r}^d\right)^2$$

---

[2] Note that computing the conditional expected value of an arbitrary function $g(y)$ of several variables may be difficult. In general $g(y)$ can be expanded as a power series to obtain $E\{g(y)|S\}$ in terms of moments of $p(y|S)$.

[3] When $S$ is just $s^s$ (there is no information about the target attributes) the constants $E_{i,r}\{g(y)\}$ can be precomputed from the model of nature $p(z)$ after the learning stage.

Therefore, given $\boldsymbol{S}$, from Tchevichev inequality we can answer $y \cong \hat{y}_{MSE}(\boldsymbol{S}) \pm 2 e_{MSE}(\boldsymbol{S})$ with a confidence level above 75%. When the shape of $p(y|\boldsymbol{S})$ is complex it must be reported explicitly (the point estimator $\boldsymbol{y} \cong \hat{\boldsymbol{y}}_{MSE}(\boldsymbol{S})$ only makes sense if $p(y|\boldsymbol{S})$ is unimodal).

<u>Example 10</u>: *Nonlinear regression*. Fig. 11 shows the mixture components and regression lines (with a confidence band of two standard deviations) obtained in a simple example of nonlinear dependence between two variables. In this case the joint density can be adequately approximated by 3 or 4 components: MSE (1 component) = 0.532, MSE (2 comp.) = 0.449, MSE (3 comp.) = 0.382, MSE (4 comp.) = 0.381.

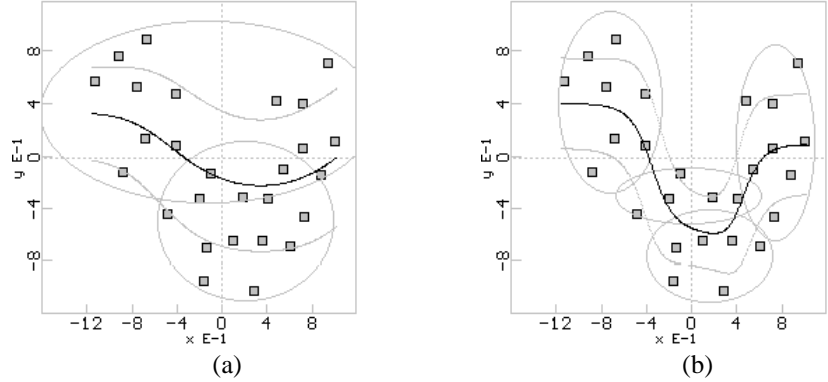

(a)             (b)

Figure 11. Nonlinear regression example: (a) 2 components, (b) 4 components.

When the target $y$ is symbolic we must compute the posterior probability of each value. In this case all the $\lambda_{i,r}^{d} = 0$ and the $v_{i,r}^{d} = \mu_{i}^{d}$ are the possible values $\omega$ taken by $y = z^{d}$. Collecting together all the $v_{i,r}^{d} = \omega$, as in (8), eq. (15) can be written as:

$$p(y|\boldsymbol{S}) = \sum_{\omega} \sum_{i,r} \alpha_{i,r,\omega} \ \mathcal{N}(y,\omega)$$

where $\alpha_{i,r,\omega}$ are the coefficients of the impulses located at $\omega$. The posterior probability of each value is:

$$q_{\omega} \equiv \mathrm{P}\{y = \omega|\boldsymbol{S}\} = \sum_{i,r} \alpha_{i,r,\omega}$$

For instance, the minimum error probability estimator (EP) is:

$$\hat{y}_{EP}(\boldsymbol{S}) = argmax_{\omega} \ q_{\omega}$$

and any desired rejection threshold can be easily established. We can reject the decision if the entropy of the posterior, $H = -\Sigma_{\omega} q_{\omega} \ log \ q_{\omega}$, or the estimated error probability, $E = 1- \ max \ q_{\omega}$, are too high.

<u>Example 11</u>: *Nonparametric Pattern Recognition*. Fig. 12 shows a bivariate data set with elements from two different categories, represented as the value of an additional symbolic attribute. The joint density can be satisfactorily approximated by a 6-component mixture (Fig. 12.a). The decision regions when the rejection threshold was set to 0.9 are shown in Fig. 12.b.

Note that Statistical Pattern Classification usually start from an (implicit or explicit) approximation to the class-conditional densities. In contrast, we start from the joint density, from which the class-conditional densities can be easily derived (Fig. 12.c).

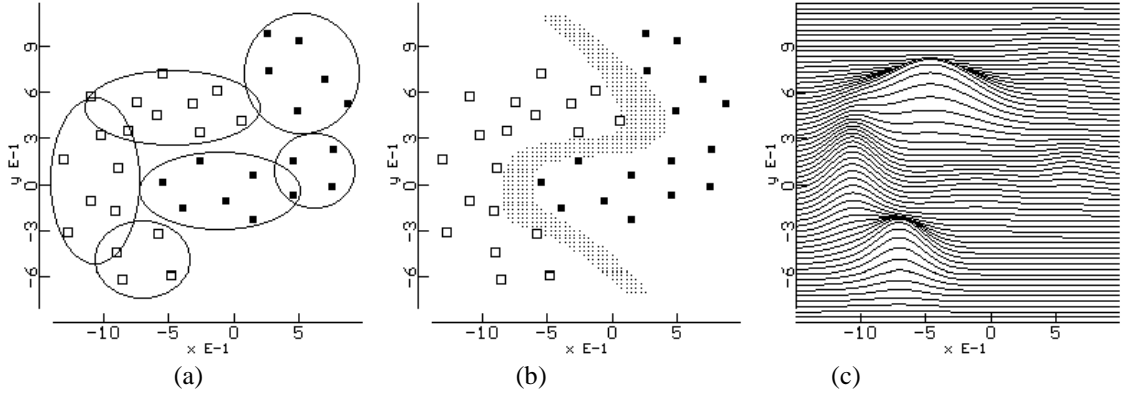

(a)      (b)      (c)

Figure 12. Simple nonparametric 2–feature pattern recognition task and its 3–attribute joint mixture model: (a) Feature space and mixture components. (b) Decision boundary. (c) One of the class-conditional densities.

The computation of the optimum estimators for other loss functions is straightforward. Observe that the estimators are based on the combination of different rules, weighted by their degree of applicability. This is a typical structure used by many other decision methods. In our case, since the components of the joint density have independent variables the rules reduce to constants, the simplest type of rule.

### 3.6  Examples of Elementary Pieces of Information

Some important types of elementary observations $s_r^j$ about $z^j$ are shown, including the corresponding likelihoods $\beta_{i,r}^j$ and modified marginals $\psi_{i,r}^j(z^j)$ ($j=d$) required in expression (15).

**Exact information**: $s_r^j = z^j$. The observation is modeled by an impulse: $p(s_r^j|z^j) = \mathcal{N}(s_r^j, z^j) = \delta(s_r^j - z^j)$. Therefore:

$$\beta_{i,r}^j = \mathcal{N}(s_r^j, \mu_i^j, \sigma_i^j)$$
$$\psi_{i,r}^j(z^j) = \mathcal{N}(z^j, s_r^j)$$

The contribution $\beta_{i,r}^j$ of exact information about the input attribute $z^j$ is the standard likelihood $p(z^j|C_i)$ of the observed value $z^j$ in each component. On the other hand, if we acquire exact information about a target attribute $z^j$ (when there is only one ($R=1$) elementary observation and $s^j = z^j$) then the inference process is trivially not required: $p(z^j|\boldsymbol{S}) = \delta(z^j - s^j)$.

**Gaussian noise** with bias $\eta_r^j$ and standard deviation $\varepsilon_r^j$: The observation is modeled by a 1-component mixture: $p(s_r^j|z^j) = \mathcal{N}(s_r^j, z^j + \eta_r^j, \varepsilon_r^j)$, which can also be expressed as a 95% confidence interval $z^j \cong s_r^j + \eta_r^j \pm 2\varepsilon_r^j$. From property (2-2):

$$\beta_{i,r}^{j} = \mathcal{N}(s_r^{j}, \mu_i^{j} - \eta_r^{j}, \sqrt{(\sigma_i^{j})^2 + (\varepsilon_r^{j})^2})$$

The effect of a noisy input $z^{j} \cong s_r^{j} \pm 2\varepsilon_r^{j}$ is equivalent to the effect of an exact input $z^{j} = s_r^{j}$ in a mixture with components of larger variance: $\sigma_i^{j} \rightarrow \sqrt{(\sigma_i^{j})^2 + (\varepsilon_r^{j})^2}$. Uncertainty spreads the effect of the observation, increasing the contribution of distant components.

Example 12: Fig. 13.a shows a simple two-attribute domain approximated by a 3-component mixture. We are interested in the marginal density of attribute $x$ given different degrees of uncertainty $\varepsilon$ in the input attribute $y \approx .4 \pm 2\varepsilon$, modeled by $p(s^{y}|y) = \mathcal{N}(s^{y}, .4, \varepsilon)$. If $\varepsilon = 0$ we have the sharpest density (A) in Fig. 13.b, providing $x \approx -.4 \pm .5$. If $\varepsilon = .25$ we obtain density (B) and $x \approx -.3 \pm .7$. Finally, if $\varepsilon = .5$ we obtain density (C) and $x \approx -.2 \pm .8$. Obviously, as the uncertainty in $y$ increases, so does the uncertainty in $x$. The expected value of $x$ moves towards more distant components, which become more likely as the probability distribution of $y$ expands. In this situation an interesting effect appears: the mode of the marginal density does not change at the same rate than the mean. Uncertainty in $y$ skews $p(x)$. This effect suggests that the optimum estimators for different loss functions are not equally robust against uncertainty.

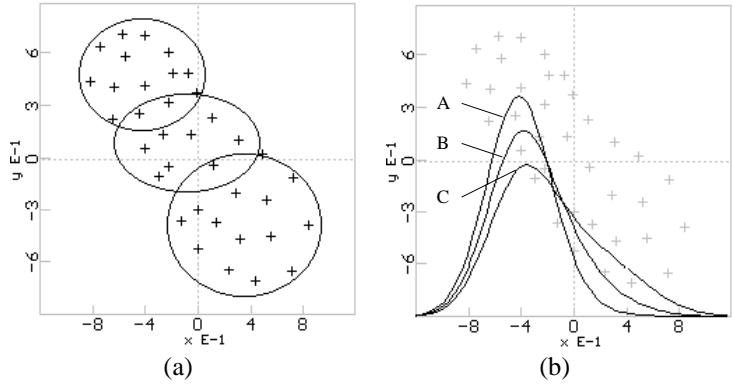

Figure 13. Effect of the amount of uncertainty (see text). (a) Data set and 3-component model. (b) $p(x \,|\, \text{uncertain } y\text{'s around } 0.4)$.

For the output role, $\psi_{i,r}^{j}(z^{j})$ becomes the original marginal, modified in location and dispersion towards $s_r^{j}$ according to the factor $\gamma = (\sigma_i^{j})^2 / [(\sigma_i^{j})^2 + (\varepsilon_r^{j})^2]$, which quantifies the relative importance of the observation:

$$\psi_{i,r}^{j}(z^{j}) = \mathcal{N}\Big(z^{j}, \gamma(s_r^{j} - \eta_r^{j}) + (1-\gamma)\mu_i^{j}, \ \gamma^{1/2}\varepsilon_r^{j}\Big)$$

**Missing data**. When there is no information about the $j$-th attribute, $s_r^{j} = \{z^{j} = ?\}$, the observation can be modeled by $p(s_r^{j}|z^{j}) = constant$ or, equivalently, $p(s_r^{j}|z^{j}) = \mathcal{N}(s_r^{j}, a, b)$ with $a$ arbitrary and $b \rightarrow \infty$. All the components contribute with the same weight:

$$\beta_{i,r}^{j} = p(z^{j} = anything|C_i) \propto constant \equiv 1$$

If the target is missing the $\psi_{i,r}^{j}(z^{j})$ reduce to the original marginal components:

$$\psi_{i,r}^j(z^j) = \mathcal{N}(z^j, \mu_i^j, \sigma_i^j) = p(z^j | C_i)$$

**Arbitrary uncertainty**. In general, any unidimensional relative likelihood function can be approximated by a mixture of generalized normals, as shown in Example 6, where $\beta_{i,r}^j$ and $\psi_{i,r}^j(z^j)$ are given respectively by eqs. (11) and (12).

**Intervals**. Some useful functions cannot be accurately approximated by a small number of normal components. A typical example is the indicator function of an interval, used to model an uncertain observation where all the values are equally likely: $s_r^j = \{z^j \in (a,b)\}$. If $z^j$ is only considered as input, we can use the shortcut $\beta_{i,r}^j = F_i^j(b) - F_i^j(a)$, where $F_i^j(z^j)$ is the cumulative distribution of the normal marginal component $p(z^j | C_i)$. Unfortunately, the expression for $\psi_{i,r}^j(z^j)$, required for $z^j$ considered as output, may not be so useful for computing certain optimum estimators. $\psi_{i,r}^j(z^j)$ is the restriction of $p(z^j | C_i)$ to the interval $(a,b)$ and normalized by $\beta_{i,r}^j$.

**Disjunction and conjunction of events**. Finally, standard probability rules are used to build structured information from simple observations: if from subjective judgments or objective evidence we ascribe relative degrees of credibility $\theta_r^j$ to several observations $s_r^j$ about $z^j$, the overall likelihood becomes $\beta_i^j = \Sigma_r \theta_r^j \beta_{i,r}^j$. In particular, if $s^j = \{z^j = \omega_1 \text{ OR } z^j = \omega_2\}$ and the two possibilities are equiprobable then $\beta_i^j = p(\omega_1 | C_i) + p(\omega_2 | C_i)$. Analogously, conjunctions of events translate to multiplication of likelihood functions.

### 3.7 Summary of the Inference Procedure

Once the domain $p(z)$ has been adequately modeled in the learning process (as explained in Section 4), the system enters the inference stage over new, partially specified objects. From the parameters of the domain $p(z)$ ($P_i$, $\mu_i^j$ and $\sigma_i^j$) and the parameters of the model of the observation $p(S/z)$ ($\pi_r$, $s_r^j$ and $\varepsilon_r^j$), we must obtain the parameters $\beta_{i,r}^j$, $\nu_{i,r}^d$ and $\lambda_{i,r}^d$ of the desired marginal posterior densities and estimators. The inference procedure comprises the following steps:

- Compute the elementary likelihoods $\beta_{i,r}^j$, using eq. (11).

- Obtain the product $\beta_{i,r}$ for each conjunction $s_r$ and component $C_i$, using eq. (14).

- Normalize $P_i \pi_r \beta_{i,r}$ to obtain the coefficients $\alpha_{i,r}$ of the posterior, using eq. (16).

- Choose the desired target attributes $y = \{z^d\}$ and compute the parameters $\nu_{i,r}^d$, and $\lambda_{i,r}^d$ of the modified component marginal densities $\psi_{i,r}^d(z^d)$ using eq. (12).

- Report the joint posterior density of $y$. Show graphs of the posterior marginal densities of the desired attributes $z^d$ using eq. (15). Provide optimum (point, interval, etc.) estimators using eq. (17).

Example 13: *Iris Data*. The inference procedure is illustrated over the well known Iris benchmark: 150 objects represented by four numeric features ($x$, $y$, $z$ and $w$) and one symbolic category $U \in \{U_1 \text{ (setosa)}, U_2 \text{ (versicolor)}, U_3 \text{ (virginica)}\}$. The whole data set was divided into two disjoints subsets for training and validation. The joint density can be satisfactorily approxi-mated (see Section 4) by a 6–component mixture (the error rate classifying $U$ in the validation set without rejection was 2.67%). Fig. 14 shows two projections of the 150 examples and the location of the mixture components learned from the training subset. The parameters of the mixture are shown in Table 3.

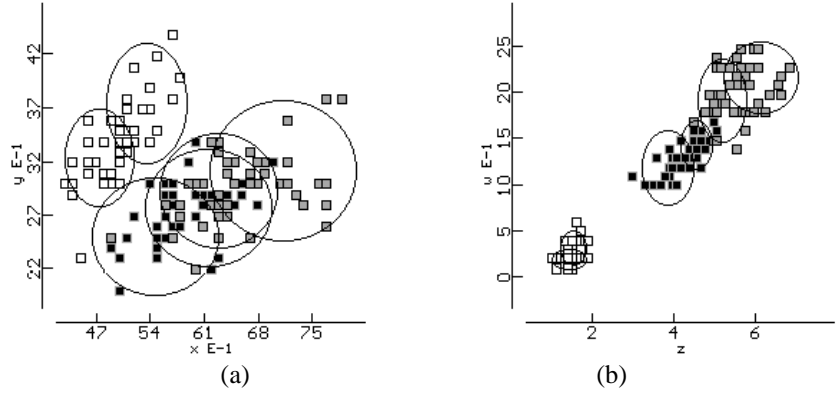

(a)                                        (b)

Figure 14. Two views of the Iris examples and the components of the joint density mixture model. $U_1$: white, $U_2$: black, $U_3$: gray. (a) Attributes $x$, $y$ (b) Attributes $z$, $w$.

| $i$ | $P_i$ | $\mu_i^x$ | $\sigma_i^x$ | $\mu_i^y$ | $\sigma_i^y$ | $\mu_i^z$ | $\sigma_i^z$ | $\mu_i^w$ | $\sigma_i^w$ | P$\{U_1|C_i\}$ | P$\{U_2|C_i\}$ | P$\{U_3|C_i\}$ |
|---|---|---|---|---|---|---|---|---|---|---|---|---|
| 1 | 0.15 | 7.13 | 0.48 | 3.12 | 0.34 | 6.17 | 0.45 | 2.18 | 0.20 | 0 | 0 | 1 |
| 2 | 0.13 | 5.48 | 0.41 | 2.50 | 0.28 | 3.87 | 0.32 | 1.20 | 0.21 | 0 | 0.93 | 0.07 |
| 3 | 0.21 | 6.29 | 0.39 | 2.93 | 0.27 | 4.59 | 0.20 | 1.45 | 0.14 | 0 | 1 | 0.00 |
| 4 | 0.18 | 4.75 | 0.23 | 3.25 | 0.23 | 1.42 | 0.21 | 0.19 | 0.05 | 1 | 0 | 0 |
| 5 | 0.15 | 5.36 | 0.26 | 3.76 | 0.29 | 1.51 | 0.16 | 0.32 | 0.10 | 1 | 0 | 0 |
| 6 | 0.17 | 6.16 | 0.42 | 2.77 | 0.28 | 5.22 | 0.30 | 1.94 | 0.23 | 0 | 0.00 | 1 |

Table 3. Parameters of the Iris Data Joint Density Model

Table 4 shows the results of the inference process in the following illustrative situations:

**Case 1**: Attribute $z$ is known: $S = \{z = 5\}$.

**Case 2**: Attributes $x$ and $U$ are known: $S = \{(x = 5.5)$ AND $(U=U_2)\}$.

**Case 3**: Attribute $x$ is uncertain: $S = \{x \cong 7\pm1\}$.

**Case 4**: Attributes $x$ and $w$ are uncertain: $S = \{(x \cong 7\pm1)$ AND $(w \cong 1\pm0.5)\}$. Note that uncertainty decreases when more information is supplied (compare with Case 3).

**Case 5**: Structured query expressed in terms of logical connectives over uncertain elementary events: $S = \{[(z \cong 1\pm3)$ OR $(z \cong 7\pm3)]$ AND $[(U = U_1)$ OR $(U = U_2)]\}$.

| CASE | | $X$ | $y$ | $z$ | $w$ | $U$ |
|------|--------|-----------|-----------|-----------|-----------|------------------------------------|
| 1 | INPUT | ? | ? | **5.0** | ? | ? |
| | OUTPUT | 6.2±0.9 | 2.8±0.6 | 5.0 | 1.8±0.6 | $U_2$: 22%   $U_3$: 78% |
| 2 | INPUT | **5.5** | ? | ? | ? | $U_2$: **100%** |
| | OUTPUT | 5.5 | 2.6±0.6 | 4.0±0.8 | 1.3±0.4 | $U_2$: 100% |
| 3 | INPUT | **7±1** | ? | ? | ? | ? |
| | OUTPUT | 6.7±0.9 | 3.0±0.7 | 5.3±1.8 | 1.8±0.8 | $U_2$: 36%   $U_3$: 63% |
| 4 | INPUT | **7±1** | ? | ? | **1±0.5** | ? |
| | OUTPUT | 6.5±0.7 | 2.9±0.6 | 4.5±0.8 | 1.3±0.3 | $U_2$: 95%   $U_3$: 5% |
| 5 | INPUT | ? | ? | **1±3** | ? | $U_1$: **50%**   $U_2$: **50%** |
| | OR | ? | ? | **7±3** | ? | $U_1$: **50%**   $U_2$: **50%** |
| | OUTPUT | 5.3±1.2 (approx. unimodal) | 3.3±0.9 (unimodal) | 2±3 (bimodal) | 0.5±1 (bimodal) | $U_1$: 75%   $U_2$: 25% |

Table 4. Some Inference Results Over the IRIS Domain

The consistency of the results can be visually checked in Fig. 14. Finally, Table 5 shows the elementary likelihoods $\beta_{i,r}^j$ of Case 5, illustrating the essence of the method.

| $i$ | $\beta_{i,1}^x$ | $\beta_{i,1}^y$ | $\beta_{i,1}^z$ | $\beta_{i,1}^w$ | $\beta_{i,1}^U$ | $\beta_{i,1}$ | $\beta_{i,2}^x$ | $\beta_{i,2}^y$ | $\beta_{i,2}^z$ | $\beta_{i,2}^w$ | $\beta_{i,2}^U$ | $\beta_{i,2}$ |
|---|---|---|---|---|---|---|---|---|---|---|---|---|
| 1 | 1 | 1 | .001 | 1 | **0** | 0 | 1 | 1 | .221 | 1 | **0** | 0 |
| 2 | 1 | 1 | .045 | 1 | **.47** | .02 | 1 | 1 | .032 | 1 | **.47** | .02 |
| 3 | 1 | 1 | .016 | 1 | **.50** | .01 | 1 | 1 | .074 | 1 | **.50** | .04 |
| 4 | 1 | 1 | .254 | 1 | **.50** | .13 | 1 | 1 | **3E-4** | 1 | **.50** | .00 |
| 5 | 1 | 1 | .250 | 1 | **.50** | .13 | 1 | 1 | **4E-4** | 1 | **.50** | .00 |
| 6 | 1 | 1 | .006 | 1 | **0** | 0 | 1 | 1 | .132 | 1 | **0** | 0 |

Table 5. Elementary likelihoods in Case 5 from Table 4.

### 3.8 Independent Measurements

One of the key features of the MFGN framework is the ability to infer over arbitrary relational knowledge about the attributes, in the form of a likelihood function adequately approximated by a mixture model with the structure of eq. (9). For instance, we could answer questions as: "what happens to $z^d$ when $z^i$ tends to be less than $z^j$?" (i.e., when $p(S|z)$ is high in the region $z^i - z^j < 0$). However, there are situations where the observations over each single attribute $z^j$ are statistically independent: we have information about attributes (e.g. $z^i$ is around $a$ and $z^j$ is around $b$) but not about attribute relations. We will pay attention to this particular case because it illustrates the role of the main MFGN framework elements. Furthermore, many practical applications can be satisfactorily solved under the assumption of independent measurements or judgments. In this case, the likelihood of the available information can be expressed as the conjunction of $n$ "marginal" observations $s^j$ about $z^j$:

$$p(S|z) = \prod_j p(s^j|z^j) \tag{18}$$

This means that the sum of products in equation (9) is "complete", i.e., it includes all the elements in the $N$-fold cartesian product of attributes:

$$p(\boldsymbol{S}|z) = \prod_j \sum_r \pi_r^j \ \mathcal{N}(z^j, s_r^j, \varepsilon_r^j)$$

where $\prod_j \pi_r^j = \pi_r$. This factored likelihood function can be considered also as a 1-component mixture (with $R=1$ in (9) and $s^j \equiv s_1^j$) where the marginal observation models are allowed to be mixtures of generalized normals: $p(s^j|z^j) = \Sigma_r \pi_r^j \mathcal{N}(z^j, s_r^j, \varepsilon_r^j)$. In this case we can even think of "function valued" attributes $\boldsymbol{z} \cong [f^1(z^1),\dots,f^n(z^n)]$, where $f^j(z^j) \equiv p(s^j|z^j)$ models the range and relative likelihood of the values of $z^j$. Loosely speaking, attributes with concentrated $f^j(z^j)$ may be considered as inputs, and attributes with high dispersion play the role of outputs. Since $\boldsymbol{y}$ is conditionally independent of $\boldsymbol{s^x}$ given $C_i$, the posterior can be obtained from the expansion:

$$p(\boldsymbol{y}|\boldsymbol{S}) = \sum_i p(\boldsymbol{y}|\boldsymbol{S}, C_i) \, \mathrm{P}\{C_i|\boldsymbol{S}\} = \sum_i p(\boldsymbol{y}|C_i, \boldsymbol{s^y}) \, \mathrm{P}\{C_i|\boldsymbol{S}\} \tag{19}$$

The interpretation of (19) is straightforward. The effect of $\boldsymbol{s^x}$ over $\boldsymbol{y} = \{z^d\}$ must be computed through $\boldsymbol{x} = \{z^o\}$ and the components $C_i$. Then, a simple Bayesian update of $p(\boldsymbol{y}|\boldsymbol{s^x})$ as a new prior is made using $\boldsymbol{s^y}$ (see Fig. 15).

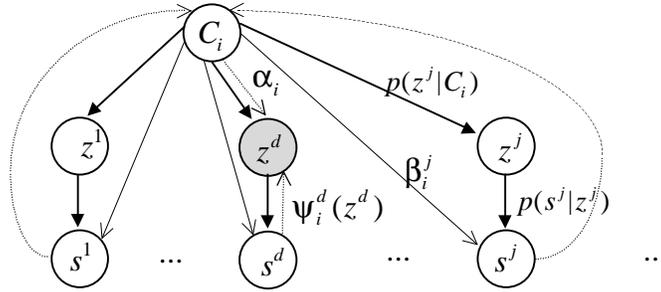

Figure 15. Structure of the MFGN inference process from independent pieces of information. In this case, the likelihood function is also factorizable. The data flow in the inference process is shown by dotted arrows.

## 4. Learning from Uncertain Information

In the previous section, we have described the inference process from uncertain information under the MFGN framework. Now we will develop a learning algorithm for the model of the domain, where the training examples will be also uncertain. Specifically, we must find the parameters $P_i$, $\mu_i^j$, $\sigma_i^j$ (or $t_{i,\omega}^j$) of a mixture with structure (7) to approximate the true joint density $p(z)$ from a training i.i.d. random sample $\{z^{(k)}\}$, $k=1..M$, partially known through the associated likelihood functions $\{S^{(k)}\}$ with structure (9).

### 4.1 Overview of the EM Algorithm

Maximum Likelihood estimates for the parameters of mixture models are usually computed by the well-known Expectation-Maximization (EM) algorithm (Dempster, Laird and Rubin 1997, Redner and Walker 1984, Tanner 1996), based in the following idea. In principle, the maximization of the training sample likelihood $J = \Pi_k p(z^{(k)})$ is a mathematically complex task due to the product of sums structure. However, note that $J$ could be conveniently expressed for maximization if the components that generated each example were known (this is called "complete data" in EM terminology). The underlying credit assignment problem disappears and the estimation task reduces to several uncoupled simple maximizations. The key idea of EM is the following: instead of maximizing the complete data likelihood (which is unknown), we can iteratively maximize its expected value given the training sample and the current mixture parameters. It can be shown that this process eventually achieves a local maximum of $J$.

Instead of a rigorous derivation of the EM algorithm, to be found in the references (see especially McLachlan and Krishnan, 1997), we will present here a more heuristic justification which provides insight for generalizing the EM algorithm to accept uncertain examples. We will review first the simplest case, where no missing or uncertain values are allowed in the training set. The parameters of the mixture are conditional expectations:

$$\mathrm{E}_{z|C_i}\{g(z)|C_i)\} = \int_Z g(z)\, p(z|C_i)\, dz \tag{20}$$

In particular, $\mu_i^j = \mathrm{E}\{z^j|C_i\}$, $(\sigma_i^j)^2 = \mathrm{E}\{(z^j - \mu_i^j)^2|C_i\}$ and $t_{i,\omega}^j = \mathrm{E}\{I\{z^j = \omega\}|C_i\}$. The mixture proportions are $P_i = \mathrm{E}\{\,\mathrm{P}\{C_i|z\}\,\}$.

We rewrite the conditional expectation (20) using Bayes Theorem in the form of an unconditional expectation:

$$\mathrm{E}_{z|C_i}\{g(z)|C_i)\} = \int_Z g(z)\, \mathrm{P}\{C_i|z\}\, p(z)\, /\, \mathrm{P}\{C_i\}\, dz\ = \tag{21}$$

$$= \mathrm{E}_z\{g(z)\, \mathrm{P}\{C_i|z\}\}\, /\, P_i \tag{22}$$

The EM algorithm can be interpreted as a method to iteratively update the mixture parameters using expression (22) in the form of an empirical average over the training data[4]. Starting from a tentative, randomly chosen set of parameters, the following E and M steps are repeated until the total likelihood $J$ no longer improves (the notation $(expression)^{(k)}$ means that $(expression)$ is computed with the parameters of example $z^{(k)}$):

(E) *Expectation step*. Compute the probabilities $q_i^{(k)} \equiv \mathrm{P}\{C_i|z^{(k)}\}$ that the $k$-th example has been generated by the $i$-th component of the mixture:

$$q_i^{(k)} \leftarrow p(z^{(k)}|C_i)\,\mathrm{P}\{C_i\}\,/\,p(z^{(k)})$$

---

[4] Expression (21) can be also used for iterative approximation of explicit functions which are not indirectly known by i.i.d. sampling (e.g., subjective likelihood functions sketched by the human user, as in Example 4). In this case $p(z)$ is set to the target function and $\mathrm{P}\{C_i|z\}$ is computed from the current mixture model.

(M) *Maximization step*. Update the parameters of each component using all the examples, weighted by their probabilities $q_i^{(k)}$. First, the *a priori* probabilities of each component:

$$P_i \leftarrow \frac{1}{M}\sum_k q_i^{(k)}$$

Then, for continuous variables, the mean values and standard deviations in each component:

$$\mu_i^j \leftarrow \frac{1}{MP_i}\sum_k [q_i\,z^j]^{(k)}$$

$$(\sigma_i^j)^2 \leftarrow \frac{1}{MP_i}\sum_k [q_i\,(z^j)^2]^{(k)} - (\mu_i^j)^2 \tag{23}$$

and for symbolic variables, the probabilities of each value:

$$t_{i,\omega}^j \leftarrow \frac{1}{MP_i}\sum_k [q_i\,I\{z^j=\omega\}]^{(k)}$$

### 4.2 Extension to Uncertain Values

In general, in the MFGN framework we do not know the true values $z^j$ of the attributes in the training examples, required to compute $g(z)\,\mathrm{P}\{C_i|z\}$ in the (empirical) expectation (22). Instead, we will start from uncertain observations $S^{(k)}$ about the true training examples $z^{(k)}$, in the form of likelihood functions expressed as mixtures of generalized normals:

$$p(S^{(k)}|z^{(k)}) = \sum_r \mathrm{P}\{S^{(k)}|s_r^{(k)}\}\,p(s_r^{(k)}|z^{(k)})$$

Therefore, we must express the expectation (22) over $p(z)$ as an unconditional expectation over $p(S)$, the distribution which generates the available information about the training set. This can be easily done by expanding $p(z|C_i)$ in terms of $S$:

$$\mathrm{E}_{z|C_i}\{g(z)|C_i)\} = \int_Z g(z)\,p(z|C_i)\,dz$$

$$= \int_Z g(z)\left[\int_S p(z|S,C_i)\,p(S|C_i)\,dS\right]dz$$

$$= \int_S \left[\int_Z g(z)\,p(z|S,C_i)\,dz\right]\mathrm{P}\{C_i|S\}\,p(S)\,dS\,/\,\mathrm{P}\{C_i\} \tag{24}$$

If we define

$$\Gamma_i(S) \equiv \mathrm{E}_{z|S,C_i}\{g(z)|S,C_i\} = \int_Z g(z)\,p(z|C_i)\,p(S|z)\,dz\,/\,p(S|C_i)$$

then the parameters of $p(z)$ can be finally written[5] as an unconditional expectation over the observable $p(S)$ in a form similar to eq. (22):

$$\mathrm{E}_{z|C_i}\{g(z)|C_i)\} = \mathrm{E}_S\{\Gamma_i(S)\,\mathrm{P}\{C_i|S\}\}\,/\,P_i \qquad (25)$$

This expression justifies an extended form of the EM algorithm to iteratively update the parameters of $p(z)$ by averaging $\Gamma_i(S)\,\mathrm{P}\{C_i|S\}$ over the available training information $\{S^{(k)}\}$ drawn from $p(S)$. This can be considered as a numerical/statistical method for solving $p(z)$ in the integral equation:

$$\int_Z p(S|z)\,p(z)\,dz = p(S)$$

Note that we cannot approximate $p(S)$ as a fixed mixture in terms of $p(S|C_i)$ and then computing back the corresponding $p(z|C_i)$ because, in general, $p(S|z)$ will be different for the different training examples. For the same reason, elementary deconvolution methods are not directly applicable.

This kind of problem is addressed by Vapnik (1982, 1995) to perform inference from the result of "indirect measurements". This is an ill-posed problem, requiring regularization techniques. The proposed extended EM algorithm can be considered as a method for empirical regularization, in which the solution is restricted to be in the family of mixtures of (generalized) gaussians. EM is also proposed by You and Kaveh (1996) for regularization in the context of image restoration.

The interpretation of (25) is straightforward. Since we do not know the exact $z$ required to approximate the parameters of $p(z)$ by empirically averaging $g(z)\,\mathrm{P}\{C_i|z\}$, we obtain the same result by averaging the corresponding $\Gamma_i(S)\,\mathrm{P}\{C_i|S\}$ in the $S$ domain, where $\Gamma_i(S)$ plays the role of $g(z)$ in (22). As $z$ is uncertain, $g(z)$ is replaced by its expected value in each component given the information about $S$. In particular, if there is exact knowledge about the training set attributes ($S^{(k)}=z^{(k)}$, i.e., $R=1$ and the marginal likelihoods are impulses) then (25) reduces to (22). Fig. 16. illustrates the approximation process performed by the extended version of the EM algorithm in a simple univariate situation.

It is convenient to develop a version of the proposed Extended EM algorithm for uncertain training sets, structured as tables of (sub)cases × (uncertainly valued) variables (see Fig. 17).

First, let us write eq. (24) expanding $S$ in terms of its components $s_r$:

$$p(z|C_i,S)\,\mathrm{P}\{C_i|S\} = p(z,C_i|S)$$

$$= \sum_r p(z,C_i|s_r)\,\mathrm{P}\{s_r|S\} = \sum_r p(z|C_i,s_r)\,\mathrm{P}\{C_i|s_r\}\,\mathrm{P}\{s_r|S\}$$

Therefore

$$\Gamma_i(S)\,\mathrm{P}\{C_i|S\} = \sum_r \Gamma_{i,r}(s_r)\,\mathrm{P}\{C_i,s_r|S\}$$

---

[5] This result can be also obtained from the relation $\mathrm{E}_z\{w(z)\} = \mathrm{E}_S\{\,\mathrm{E}_{z|S}\{w(z)|S\}\,\}$ for $w(z) \equiv g(z)\,\mathrm{P}\{C_i|z\}$ and Bayes Theorem.

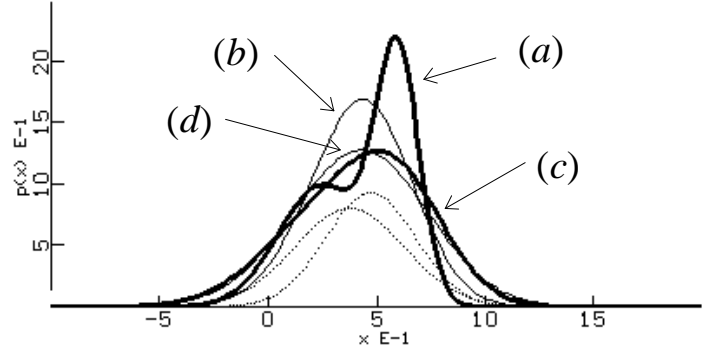

Figure 16. The extended EM algorithm iteratively reduces the (large) difference between (*a*) the true density $p(z)$, and (*b*) the mixture model $\hat{p}(z)$, indirectly through the (small) discrepancies between (*c*) the true observation density $p(\mathbf{S})$ and (*d*) the modeled observation density $\hat{p}(\mathbf{S})$. In real cases $p(\mathbf{S})$ must be estimated from a finite i.i.d. sample $\{\mathbf{S}^{(k)}\}$.

| $\mathbf{S}^{(1)}$ | .4 | $\mathbf{s}_1^{(1)}$ | | | |
| | .6 | $\mathbf{s}_2^{(1)}$ | | | |
| $\mathbf{S}^{(2)}$ | 1 | $\mathbf{s}^{(2)}$ | | | |
| $\mathbf{S}^{(3)}$ | .2 | $\mathbf{s}_1^{(3)}$ | | | |
| | .5 | $\mathbf{s}_2^{(3)}$ | | | |
| | .3 | $\mathbf{s}_3^{(3)}$ | | | |
| | ... | | | | |
| $\mathbf{S}^{(r)}$ | $\pi_r^{(k)}$ | $(\mathbf{s}_r^1, \boldsymbol{\varepsilon}_r^1)^{(k)}$ | ... | $(\mathbf{s}_r^j, \boldsymbol{\varepsilon}_r^j)^{(k)}$ | ... |
| | ... | | | | |

Figure 17. Structure of the uncertain training information for the Extended EM Algorithm. The coefficients $\pi_r^{(k)}$ are normalized for easy detection of the rows included in each uncertain example. When $\mathbf{z}^{(k)}$ is not uncertain, $\mathbf{S}^{(k)}$ reduces to a single row with $\pi^{(k)} = 1$ and all the $\boldsymbol{\varepsilon}^j = 0$.

Using the notation introduced in (12),

$$\Gamma_{i,r}(\mathbf{s}_r) \equiv \mathrm{E}_{z|\mathbf{s}_r, C_i}\{g(z)|\mathbf{s}_r, C_i\} = \int_Z g(z)\, p(z|\mathbf{s}_r, C_i)\, dz = \int_Z g(z) \prod_j \psi_{i,r}^j(z^j)\, dz$$

$$\mathrm{P}\{C_i, \mathbf{s}_r | \mathbf{S}\} = \mathrm{P}\{C_i | \mathbf{s}_r\}\, \mathrm{P}\{\mathbf{s}_r | \mathbf{S}\} = \alpha_{i,r}$$

we can write (25) as:

$$E_{\mathbf{z}|C_i}\{g(\mathbf{z}) \mid C_i\} = E_S\left\{\sum_r \alpha_{i,r} \int_Z g(\mathbf{z}) \prod_j \psi_{i,r}^j(z^j) d\mathbf{z}\right\}\Big/ P_i$$

In the MFGN framework the contributions $\Gamma_{i,r}(\mathbf{s}_r)\, P\{C_i, \mathbf{s}_r | \mathbf{S}\}$ to the empirical expected values required by the Extended EM algorithm can be obtained again without numeric integration. We only need to consider the case $g(z) = z^j$ to compute the means $\mu_i^j$ and probabilities $t_{i,\omega}^j$, and $g(z) = (z^j)^2$ for the deviations $\sigma_i^j$. From (12) we already know an explicit expression for the parameters of $\psi_{i,r}^j(z^j) = \mathcal{N}(z^j, \nu_{i,r}^j, \lambda_{i,r}^j)$. Hence:

$$\int_Z z^j \Psi_{i,r}(z)\, dz = \int_Z z^j \psi_{i,r}^j(z^j) dz^j = \nu_{i,r}^j$$

$$\int_Z (z^j)^2\, \Psi_{i,r}(z)\, dz = (\nu_{i,r}^j)^2 + (\lambda_{i,r}^j)^2$$

In conclusion, the steps of the Extended EM algorithm are as follows:

(E) *Expectation step.* Compute all the elementary likelihoods of the training set:

$$\beta_{i,r}^{j(k)} = \mathcal{N}\left(s_r^j, \mu_i^j, \sqrt{(\sigma_i^j)^2 + (\varepsilon_r^j)^2}\right)^{(k)} \tag{26}$$

Obtain the likelihood of each conjunction $\mathbf{s}_r^{(k)}$ of example $\mathbf{S}^{(k)}$ in component $C_i$:

$$\beta_{i,r}^{(k)} = \prod_j \beta_{i,r}^{j(k)}$$

Obtain the total likelihood of example $\mathbf{S}^{(k)}$:

$$\beta^{(k)} \equiv p(\mathbf{S}^{(k)}) = \sum_i \sum_r P_i\, \pi_r^{(k)}\, \beta_{i,r}^{(k)}$$

Compute the probabilities $q_{i,r}^{(k)} \equiv P\{C_i, \mathbf{s}_r^{(k)} | \mathbf{S}^{(k)}\}$ that the $r$-th component of the $k$-th example has been generated by the $i$-th component of the mixture:

$$q_{i,r}^{(k)} \leftarrow \alpha_{i,r}^{(k)} = P_i\, \pi_r^{(k)}\, \beta_{i,r}^{(k)} / \beta^{(k)}$$

(M) *Maximization step.* Update the parameters of each component $C_i$ using all the components $\mathbf{s}_r^{(k)}$ of all the examples weighted with their probabilities $q_{i,r}^{(k)}$. First, the prior probabilities of each component:

$$P_i \leftarrow \frac{1}{M} \sum_k \sum_r q_{i,r}^{(k)}$$

Then the mean value and standard deviation in each component:

$$\mu_i^j \leftarrow \frac{1}{MP_i} \sum_k \sum_r \left[q_{i,r}\, \nu_{i,r}^j\right]^{(k)}$$

$$(\sigma_i^j)^2 \leftarrow \frac{1}{MP_i} \sum_k \sum_r \left[ q_{i,r}[(\nu_{i,r}^j)^2 + (\lambda_{i,r}^j)^2] \right]^{(k)} - (\mu_i^j)^2 \tag{27}$$

For symbolic variables under representation (8) we may use:

$$\beta_{i,r}^{j(k)} = \sum_\omega P\{s_r^j = \omega\}^{(k)} t_{i,\omega}^j \tag{26}'$$

$$t_{i,\omega}^j \leftarrow \frac{1}{MP_i} \sum_k \sum_r \left[ q_{i,r} P\{s_r^j = \omega\} t_{i,\omega}^j / \beta_{i,r}^j \right]^{(k)} \tag{27}'$$

Consider the particular case in which the attributes in the training examples are contaminated with unbiased Gaussian noise. The likelihood of the uncertain observations is modeled by 1-component mixtures: $p(s^{(k)}|z^{(k)}) = \Pi_j \mathcal{N}(z^{j(k)}, s^{j(k)}, \varepsilon^{j(k)})$, where $(\varepsilon^{j(k)})^2$ is the variance of the measurement process over $z^{j(k)}$ which obtains the observed value $s^{j(k)}$. This can be also expressed as a confidence interval $z^{j(k)} \cong s^{j(k)} \pm 2\varepsilon^{j(k)}$. In this case, the basic EM algorithm (23) can be easily modified to take into account the effect of the uncertainties $\varepsilon^{j(k)}$. In the E step, compute $q_i^{(k)}$ using the following deviations:

$$\sigma_i^j \leftarrow \sqrt{(\sigma_i^j)^2 + (\varepsilon^{j(k)})^2}$$

and, in the M step, apply the substitution:

$$z^{j(k)} \leftarrow \gamma \, s^{j(k)} + (1 - \gamma) \, \mu_i^j$$

$$(z^{j(k)})^2 \leftarrow \left[ \gamma s^{j(k)} + (1 - \gamma)\mu_i^j \right]^2 + \gamma \left[ \varepsilon^{j(k)} \right]^2$$

where

$$\gamma = \frac{(\sigma_i^j)^2}{(\sigma_i^j)^2 + (\varepsilon^{j(k)})^2}$$

measures the relative importance of the observed $s^{j(k)}$ for computing the new $\mu_i^j$ and $\sigma_i^j$.

The previous situation illustrates how missing values must be processed in the learning stage. If $z^{j(k)}$ is exact then $\varepsilon^{j(k)} = 0$ and $\gamma = 1$, so the original algorithm (23) is not changed. In the other extreme, if $z^{j(k)}$ is missing, which can be modeled by $\varepsilon^{j(k)} \equiv \infty$, we get $\gamma = 0$ and therefore the observation $s^{j(k)}$ does not contribute to the new parameters at all. The correct procedure to deal with missing values *in the MFGN framework* is simply omitting them in the empirical averages. Note that this fact arises from the factorized structure of the mixture components, providing conditionally independent attributes. Alternative learning methods require a careful management of missing data to avoid biased estimators (Ghahramani & Jordan 1994).

### 4.3 Evaluation of the Extended EM Algorithm

We have studied the improvement of the parameter estimations when the uncertainty of the observations, modeled by likelihood functions, is explicitly taken into account. The proposed Extended

EM is compared with the EM algorithm over the "raw" observations (Basic EM), which ignores the likelihood function and typically uses just its average value (e.g., given $x \cong 8 \pm 2$, Basic EM uses $x$=8). We considered a synthetic 3-attribute domain with the following joint density:

$$p(x,y,w) = 0.5 \ \mathcal{N}(x,0,2) \ \mathcal{N}(y,0,1) \ \mathcal{N}(w,white)$$
$$+ \ 0.5 \ \mathcal{N}(x,2,1) \ \mathcal{N}(y,2,2) \ \mathcal{N}(w,black)$$

Different learning experiments were performed with varying degrees of uncertainty. In all cases the training sample size was 250. All trained models had the same structure as the true density (2 components), since the goal of this experiment is to measure the quality of the estimation with respect to the amount of uncertainty, without regard of other sources of variability such as local minima, alternative solutions, etc., which are empirically studied in Section 5. Table 6 shows the mixture parameters obtained by the learning algorithms. Fig. 18 graphically shows the difference between Extended and Basic EM in some illustrative cases.

**Case 0**: Exact Data (Fig. 18.a).

**Cases *m* %**: Results of the Extended EM learning algorithm when there is a *m* % rate of missing values in the training data.

**Case 1**: Basic EM when attribute $y$ is biased +3 units with probability 0.7. **Case 2**: Extended EM algorithm over Case 1 (see Fig. 18.b). Here, the observed value is $s_y$=$y$+3 in 70% of the samples and $s_y$=$y$ in the rest. In all samples, Basic EM uses the observed value $s_y$ and Extended EM uses the explicit likelihood function $f(y) = 0.3 \ \delta(y - s_y) + 0.7 \ \delta(y - (s_y - 3))$.

**Case 3**: Basic EM when attributes $x$ and $y$ have Gaussian noise with $\sigma = 0.5$ and $w$ is changed with probability 0.1. **Case 4**: Extended EM algorithm over Case 3.

**Case 5**: Basic EM when $x$ and $y$ have Gaussian noise with $\sigma = 1$ and $w$ is changed with probability 0.2. **Case 6**: Extended EM algorithm over Case 5 (see Fig. 18.c).

**Case 7**: Basic EM when $x$ and $y$ have Gaussian noise with $\sigma = 2$ and $w$ is changed with probability 0.3. **Case 8**: Extended EM algorithm over Case 7 (see Fig. 18.d).

**Case 9**: Extended EM when values $y$>3 are missing (*censoring*). **Case 10**: Extended EM over Case 9 when the missing $y$ values are assumed to be distributed as $\mathcal{N}(y, 4, 1)$, providing some additional information on the data generation mechanism.

Table 6 and Fig. 18 confirm that for small amounts of deterioration in relation to the sample size, the estimates computed by the basic EM Algorithm over the "raw" observed data are similar to those obtained by the Extended EM algorithm (e.g., Cases 3 and 4). However, when the data sets are moderately deteriorated the true joint density can be correctly recovered by Extended EM using the likelihood functions of the attributes instead of the raw observed data (e.g., Cases 5 and 6, Fig. 18.c). Finally, when there is a very large amount of uncertainty with respect to the training sample size the true joint density cannot be adequately recovered (e.g., Cases 7 and 8, Fig. 18.d).

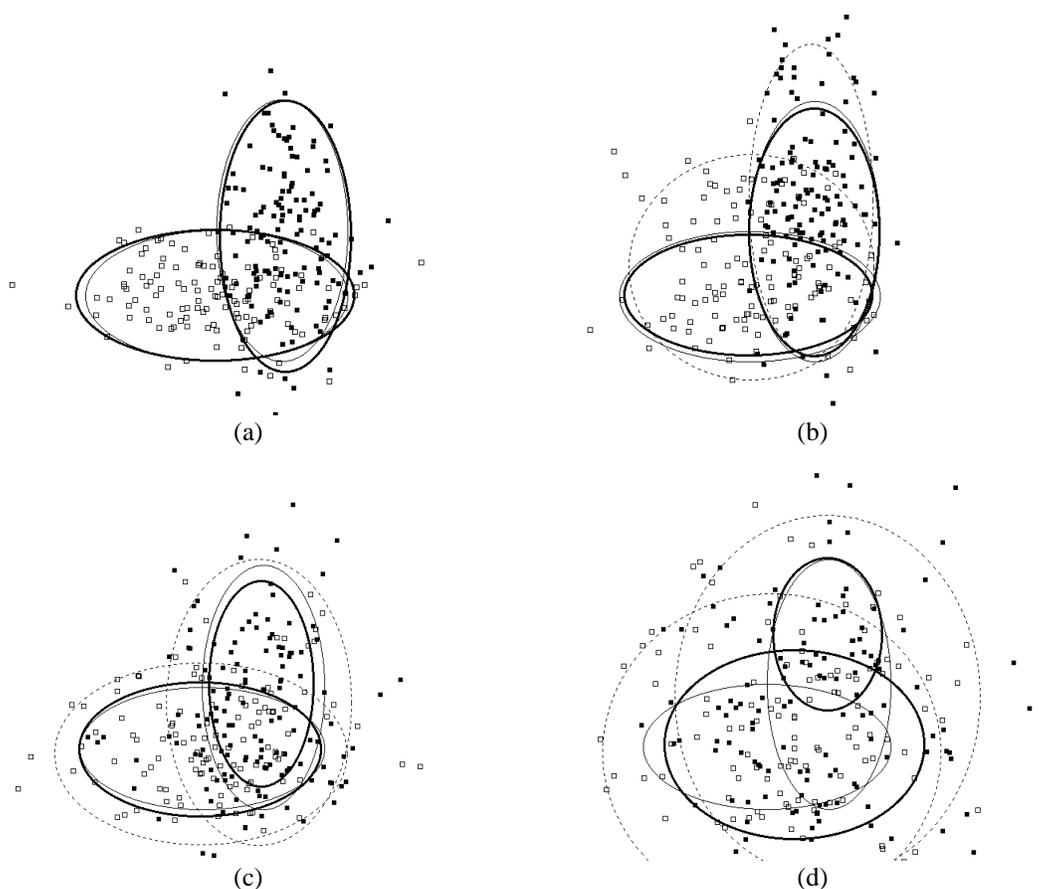

Figure 18. Illustration of the advantages of the Extended EM algorithm (see text). (a) Case 0 (exact data). (b) Cases 1 and 2 (biased data). (c) Cases 5 and 6 (moderated noise). (d) Cases 7 and 8 (large noise). All figures show the true mixture components (gray ellipses), the available raw observations (black and white squares), the components estimated by Basic EM from the raw observations (dotted ellipses) and the components estimated by Extended EM taking into account the likelihood functions of the uncertain values (black ellipses).

Note that the ability to learn from uncertain information suggests a method to manage non random missing attributes (e.g., censoring) (Ghahramani & Jordan 1994) and other complex mechanisms of uncertain data generation. As illustrated in Case 9, if the missing data generation mechanism depends on the value of the hidden attribute, it is not correct to assign equal likelihood to all components. In principle, statistical studies or other kind of knowledge may help to ascertain the likelihood of the true values as a function of the available observations. For instance, in Case 10 we replaced the missing attributes of Case 9 by normal likelihoods $y \cong 4 \pm 2$ (i.e, "$y$ is high"), improving the estimates of the mixture parameters.

| Case | $P_1$ | $\mu_1^x$ | $\mu_1^y$ | $\sigma_1^x$ | $\sigma_1^y$ | $t_{1,white}^w$ | $\mu_2^x$ | $\mu_2^y$ | $\sigma_2^x$ | $\sigma_2^y$ | $t_{2,black}^w$ |
|---|---|---|---|---|---|---|---|---|---|---|---|
| true | .5 | 0 | 0 | 2 | 1 | 1 | 2 | 2 | 1 | 2 | 1 |
| 0 | .48 | −.04 | .03 | 2.11 | 1.00 | 1.00 | 2.09 | 1.83 | 1.00 | 2.08 | 1.00 |
| 20% | .48 | .16 | −.03 | 1.91 | 1.01 | 0.96 | 2.31 | 2.09 | .88 | 2.10 | 1.00 |
| 40% | .49 | .14 | −.16 | 1.78 | .99 | 0.95 | 2.39 | 2.29 | .94 | 2.14 | 1.00 |
| 60% | .45 | .02 | −.29 | 2.50 | .78 | 1.00 | 1.86 | 2.01 | 1.03 | 1.77 | 1.00 |
| 80% | .49 | −.11 | 1.69 | 2.21 | 1.73 | .50 | 1.91 | 0.31 | 1.14 | 0.68 | 0.47 |
| 1 | .48 | .06 | .90 | 1.88 | 1.73 | 1.00 | 1.88 | 2.98 | .96 | 2.40 | 1.00 |
| 2 | .48 | −.04 | .05 | 1.90 | .92 | 1.00 | 1.98 | 1.97 | 1.01 | 1.90 | 1.00 |
| 3 | .47 | .02 | .09 | 1.60 | 1.02 | .87 | 2.00 | 1.78 | 1.14 | 2.07 | 0.85 |
| 4 | .49 | .27 | −.08 | 1.97 | .90 | .70 | 2.10 | 2.06 | 1.01 | 1.94 | 0.71 |
| 5 | .43 | −.04 | −.18 | 2.40 | 1.48 | .82 | 1.85 | 1.52 | 1.52 | 2.33 | .80 |
| 6 | .54 | −.07 | −.02 | 1.97 | 1.09 | .56 | 1.93 | 2.11 | .85 | 1.69 | .62 |
| 7 | .46 | .15 | −.16 | 2.73 | 2.53 | .31 | 1.94 | 1.62 | 2.47 | 2.90 | .29 |
| 8 | .79 | .87 | .08 | 2.09 | 1.52 | .51 | 1.96 | 3.61 | 0.87 | 1.21 | .54 |
| 9 | .48 | .32 | −.02 | 1.77 | 1.10 | 1.00 | 1.92 | 0.67 | 0.94 | 1.22 | 1.00 |
| 10 | .45 | .00 | .03 | 2.20 | 1.01 | 1.00 | 2.13 | 1.55 | 1.04 | 1.77 | 1.00 |

Table 6. Parameter Estimates from Uncertain Information (see text)

Example 14: Learning from examples with missing attributes has been performed over the IRIS domain to illustrate the behavior of the MFGN framework. The whole data set was randomly divided into two subsets of equal size for training and testing. 5-component mixture models were obtained and evaluated, combining missing data proportions of 0% and 50%. The error prediction on attribute U (*plant class*) was the following:

|  | training set | test set | prediction error |
|---|---|---|---|
| missing attributes | 0% | 0% | 2.7% |
|  | 0% | 50% | 12.0% |
|  | 50% | 0% | 4.0% |
|  | 50% | 50% | 18.7% |

In the relatively simple IRIS domain, the performance degradation due to 50% missing attributes is much greater in inference than in learning stage. The Extended EM algorithm is able to correctly recover the overall structure of the domain from the available information.

## 4.4 Comments

Convergence of the EM Algorithm is very fast, requiring no adjustable parameters such as learning rates. The algorithm is robust with respect to the random initial mixture parameters: bad local maxima are not frequent and alternative solutions are usually equally acceptable. All the examples contribute to all the components, which are never wasted by unfortunate initialization. For a fixed number of components, the algorithm progressively increases the likelihood $J$ of the training data until a maximum is reached. When the number of components is incremented the maximum $J$ also increases, until a limit value is obtained that cannot be improved using extra components (Fukunaga 1990). Some simple heuristics can be incorporated to the standard Expectation-Maximization scheme to control the value of certain parameters (e.g., lower bounds can be established for variances) or the quality of the model (e.g., mixture components can be eliminated if their proportions are too small).

In our case, factorized components are specially convenient because matrix inversions are not required and, what is more important, uncertain and missing values can be correctly handled in a simple and unified way, for heterogeneous attribute sets. It is not necessary to provide models for uncertain attribute correlations since no covariance parameters must be estimated. Finally, the training sample size must be large enough in relation to both the degree of uncertainty of the examples and the complexity of the joint density model in order to obtain satisfactory approximations.

On the other hand, the number of mixture components required for a satisfactory approximation to the joint density must be specified. A pragmatic option is the minimization of the experimental estimation cost over the main inference task, if it exists. For instance, in regression we could increase the number of components until an acceptable estimation error is obtained over an independent data set (cross-validation). The same idea applies to pattern classification: use the number of components that minimizes the error rate over an independent test set. However, one of the main advantages of the proposed method is the independence between the learning stage and the inference stage, where we can freely choose and dynamically modify the input and output role of the attributes. Therefore, a global validity criterion is desirable. Some typical validation methods for mixture models are reviewed in McLachlan & Basford (1988); the standard approach is based on likelihood ratio tests on the number of components. Unfortunately, this method does not validate the mixture itself, only selects the best number of components (DeSoete 1993).

Since the MFGN framework provides an explicit expression for the model $p(z)$, we can apply statistical tests of hypothesis over an independent sample $T$ taken from the true density (e.g. a subset of the examples reserved for testing) to find out if the obtained approximation is compatible with test data. If the hypothesis $H = \{T$ comes from $p(z)\}$ is rejected, then the learning process must continue, possibly increasing the number of components. It is not difficult to build some statistical tests, e.g. over moments of $p(z)$, because their sample means and variances can be directly obtained. However, as data sets usually include symbolic and numeric variables, we have also developed a test on the expected *likelihood* of the test sample, which measures how well $p(z)$ "covers" the examples. The mean and variance of $p(z)$ can be easily obtained using the properties of generalized normals. Some experiments over simple univariate continuous densities show that this test is not very powerful for small sample sizes, i.e. incompatibility is not always detected, while other standard tests significantly evidence rejection. Nevertheless, clearly inaccurate approximations are detected, results improve as the sample size increases and the test is valid for data sets with uncertain values.

The Minimum Description Length (Li & Vitànyi 1993) principle can be also invoked to select the optimum number of components by trading-off the complexity of the model and the accuracy in the description of the data.

## 5. Discussion and Experimental Results

### 5.1  Advantages of Joint Models

Most inductive inference methods compute a direct approximation of the conditional densities of interest, or even obtain empirical decision rules without explicit models of the underlying conditional densities. In these cases, both the model and the learning stage depend on the selected input / output role of the variables. In contrast, we have presented an inference and learning method based on an approximation to the joint probability density function of the attributes by a convenient parametric family (a special mixture model). The MFGN framework works as a pattern completion machine operating over possibly uncertain information. For example, given a pattern classification problem, the same learning stage suffices for predicting class labels from feature vectors and for estimating the value of missing features from the observed information in incomplete patterns. The joint density approach finds the regions occupied by the training examples in the whole attribute space. The attribute dependences are captured at a higher abstraction level than the one provided by strictly empirical rules for pre-established target variables. This property is extremely useful in many situations, as shown in the following examples.

Example 15: *Hints* can be provided for inference over multivalued relations. Given the data set and model from Example 10, assume that we are interested in the value of $x$ for $y = 0$. We obtain the bimodal marginal density shown in Fig. 19.a and the corresponding estimator $x \cong 0.2 \pm 1.4$ which is, in some sense, meaningless. However, if we specify the *branch* of interest of the model, inferring from $y = 0$ AND $x \cong -1\pm1$ (i.e., "$x$ is small"), we obtain the unimodal marginal density in Fig. 19.b and the reasonable estimator $x \cong -0.8\pm0.5$.

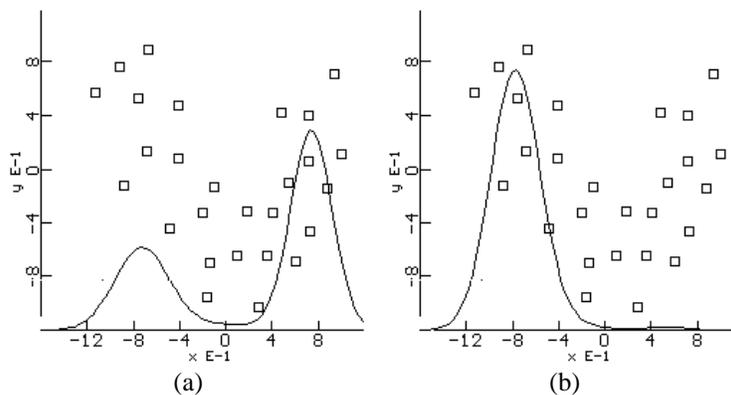

(a)                                          (b)

Figure 19. The desired branch in multivalued relations can be selected by providing some information about the output values. (a) Bimodal posterior density inferred from $y=0$. (b) Unimodal posterior density inferred from $y = 0$ and the hint "$x$ is small".

Example 16: *Image Processing*. The advantages of a joint model supporting inferences from partial information on both inputs and outputs can be illustrated in the following application with natural data (see Fig. 20). The image in Fig. 20.a is characterized by a 5-attribute density $(x, y, R, G, B)$ describing position and color of the pixels. A random sample of 5000 pixels was used to build a 100-component mixture model. We are interested in the location of certain ob-jects in the image. Figs. 20.b-f show the posterior density of the desired attributes given the following queries:

- "Something light green". Fig. 20.b. Two groups can be easily identified in the posterior density, corresponding to the upper faces of the green objects[6]. $S = C_1 = \{x, y$ unknown; $R=110\pm50$, $G=245\pm10$, $B=160\pm50\}$.

- "Something light green OR dark red". Fig. 20.c. We find the same groups as above and an additional, more scattered group, corresponding to the red object. This greater dispersion arises from the larger size of the red object and also from the fact that the R component of dark red is more disperse than the G component of light green. $S =$ Two equiprobable components with $C_1$ as above and $C_2 = \{x, y$ unknown; $R=110\pm10$, $G=B=30\pm50\}$.

- "Something light green on the right". Fig. 20.d. Here we provide partial information on the output: $S = C_3 = \{x=240\pm30$; $y$ unknown; $R=110\pm50$, $G=245\pm10$, $B=160\pm50\}$

- "Something white". Fig. 20.e. $S = C_4 = \{x,y$ unknown; $R=245\pm10$, $G=245\pm10$, $B=245\pm10\}$

- "Something white, in the lower-left region, under the main diagonal ($y<240$-$x$)". Fig. 20.f. Here we provide relational information on the attributes that can be modeled by $S = 6$ equiprobable components (note that in this case the posterior distribution contains 600 components, but it is still computationally manageable) =

$$\{x=60\pm30,\ y=180\pm30,\ R=245\pm10,\ G=245\pm10,\ B=245\pm10\}+$$
$$\{x=60\pm30,\ y=120\pm30,\ R=245\pm10,\ G=245\pm10,\ B=245\pm10\}+$$
$$\{x=60\pm30,\ y=60\pm30,\ R=245\pm10,\ G=245\pm10,\ B=245\pm10\}+$$
$$\{x=120\pm30,\ y=120\pm30,\ R=245\pm10,\ G=245\pm10,\ B=245\pm10\}+$$
$$\{x=120\pm30,\ y=60\pm30,\ R=245\pm10,\ G=245\pm10,\ B=245\pm10\}+$$
$$\{x=180\pm30,\ y=60\pm30,\ R=245\pm10,\ G=245\pm10,\ B=245\pm10\}$$

In all cases, the posterior density is consistent with the structure of the original image. The time required to compute the posterior distribution is always lower than one second. Learning time was of order of hours in a Pentium 100 system. Simpler models (25-component, obtained from 1000 random pixels) produced also acceptable results with much lower learning time. Furthermore, the EM algorithm can be efficiently parallelized.

On the other hand, when there is a large number of irrelevant attributes, the joint model strategy wastes resources to capture a proper probability density function along unnecessary dimensions. (This problem does not arise in the specification of a likelihood function, since only the relevant attributes explicitly appear in the model.) Joint modeling is appropriate for domains with a moderated number of "meaningful" variables without fixed input / output roles.

---

[6] Note that a sharp peak (a component with small dispersion) was obtained in the learning process, which also "transmits" to the posterior density. This kind of artifacts are inocuous and can be easily removed by post processing.

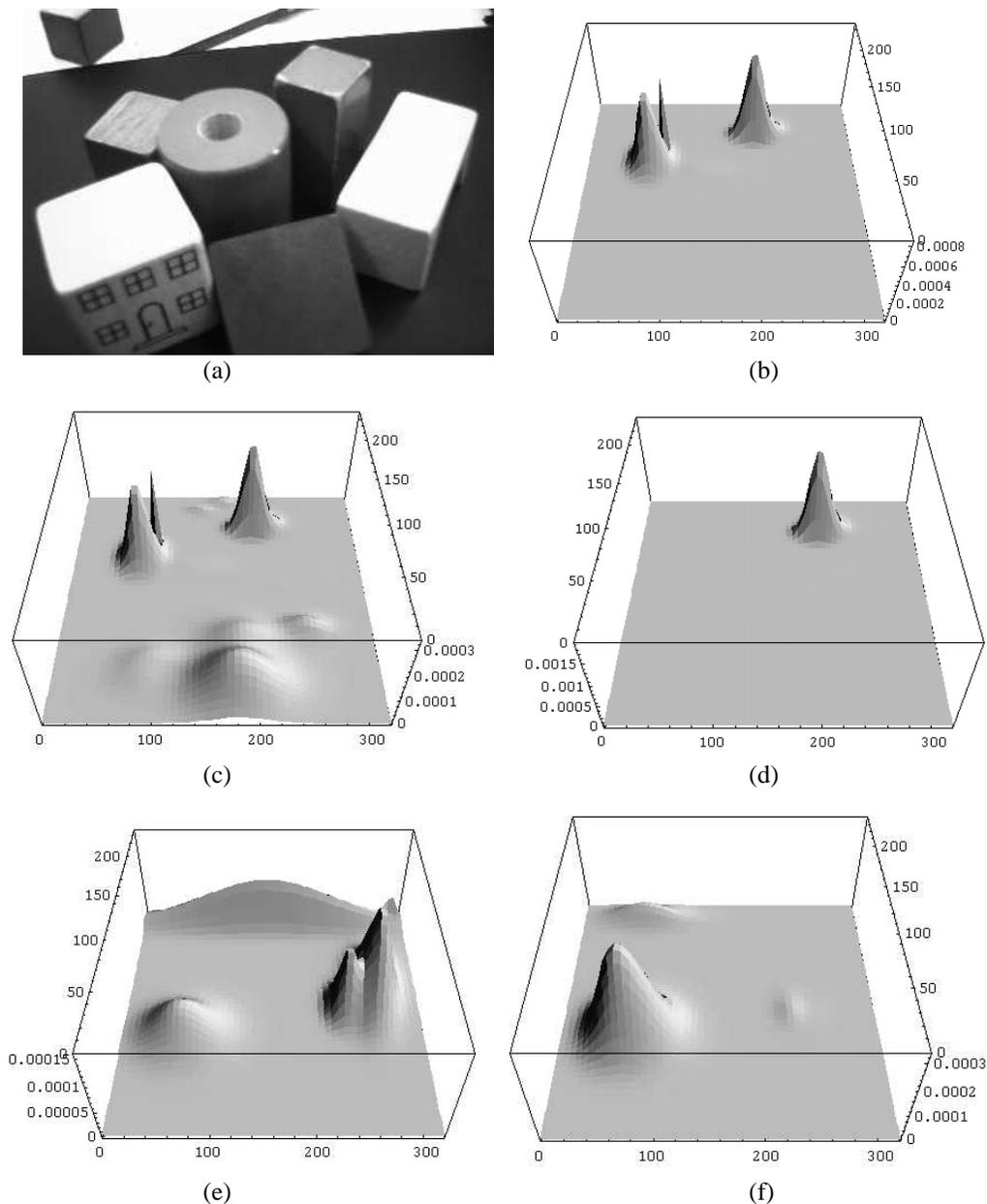

Figure 20. Inference results for the image domain in Example 16. (a) source image.
(b) posterior density of attributes *x-y* given "Something light green". (c) the same for
"Something light green OR dark red". (d) for "Something light green on the right".
(e) for "Something white". (f) for "Something white, in the lower-left region, under the
main diagonal of the image (Y<240-X)"

## 5.2 Advantages of Factorization

The proposed methodology is supported by the general density approximation property of mixture
models. We use components with independent variables in order to make computations feasible in the inference and learning stage. Factorized components can be imposed to a mixture model without loss of generality. Any statistical dependence between variables can be still captured, at the cost of a possibly larger number of components in the mixture to achieve the required accuracy in the approximation.

The simplicity of the "building block" structure is entirely compensated by an important saving in computation time. High-dimensional integrals are analytically computed from univariate integrals and matrix inversions are avoided in the learning stage. Additionally, high-dimensional domains can be easily modeled using a small number of parameters in each mixture component. From the viewpoint of Computational Learning Theory (Vapnik 1995), models with a small number of adjustable parameters (actually, with low "expressive power") have favorable consequences for generalization.

Mixtures of factorized components are also used in Latent Class Analysis (DeSoete 1993), a well-known unsupervised classification technique. It is assumed that the statistical dependences between attributes can be fully explained by a hidden variable specifying the "latent class" of each example. This method is similar to the Gaussian decomposition clustering algorithm mentioned in Section 1, constrained to component-conditional attribute independence. However, our goal is not unsupervised classification but obtaining an accurate and mathematically convenient expression for the joint density of the variables, required to derive the desired estimators. The meaning of the components is irrelevant, as long as the whole mixture is a good approximation to the joint density.

More expressive architectures, which combine mixture models with local dimensionality reduction, have been also considered: Mixtures of Linear Experts (Jordan & Jacobs 1994), Mixtures of Principal Component Analyzers (Sung & Poggio 1998) or Mixtures of Factor Analyzers (Ghahramani & Hinton 1996, Hinton, Dayan, & Revow 1997). Unfortunately, the general kind of inference and learning from uncertain data considered in this work cannot be directly incorporated into these architectures with the computational advantages demonstrated by the MFGN model.

The restriction to factorized components may produce undesirable artifacts in the approximations of certain domains learned from small training samples. Nevertheless, this problem always occurs to any approximator when the structure of the building block does not match the "shape" of the target function. In this case, many terms (or components, units, etc.) are required for a good approximation and the associated parameters can be correctly adjusted only from a large training sample. However, note that the complexity of the model should not be measured uniquely in terms of the number of mixture components. The number of adjustable parameters is probably a better measure of complexity. For instance, full covariance models show a quadratic growth of the number of free parameters with respect to the dimension of the attribute vector. For factorized components the growth is linear, so the amount of training data need not be unreasonably high even if the number of mixture components is large.

In real applications, the nature of the target function is unknown, so little can be said *a priori* about the best building block structure to be used by a universal approximator. We have chosen a very simple component structure to make inference and learning feasible from uncertain information. Section 5.4 provides experimental evidence that in realistic problems the proposed model is not inferior to other popular approaches.

### 5.3 Qualitative Comparison with Alternative Approaches

Instead of the proposed methodology, based on mixture models and the EM algorithm, other alternative nonparametric density approximation methods could also be used (either for the joint density or for specific conditional densities). For instance, the nearest neighbor rule locally approximates the target density using a certain number of training samples near to the point of interest. Symbolic attributes are directly estimated by a voting scheme and continuous attributes can be also estimated by averaging the observed values of training instances which are near, in the subspace of observed attributes, to the point of interest. However, for small sample sizes, the above estimators are not smooth and show strong sensitivity to random fluctuations in the training set, which penalizes the estimation cost. For large sample size, the time required to find the nearest neighbors becomes very long. As an example, consider the regression problem in Example 10, Section 3.5. Fig. 11.b shows the MFGN solution with 4 components and MSE=0.381. Fig. 21.a shows the regression line obtained by 5-nearest-neighbors average, with a higher MSE=0.522.

Parzen windows and similar kernel approximation methods are used to smooth the results of the simple nearest neighbors rule (Duda & Hart 1973, Izenman 1991). They are actually mixtures of simple conventional densities located at the training samples. In principle, the properties of the MFGN framework could be adapted to that kind of approximation (Ruiz et al. 1998). Learning becomes trivial, but strong run time computation effort is required since a "concise" model of the domain is not extracted from the training set. This kind of rote learning has also negative consequences on generalization according to the Occam Razor Principle (Li & Vitànyi 1993). An adequately cross-validated mixture model with a small number of components in relation to the training sample size reasonably guarantees that probably the true attribute dependencies are correctly captured.

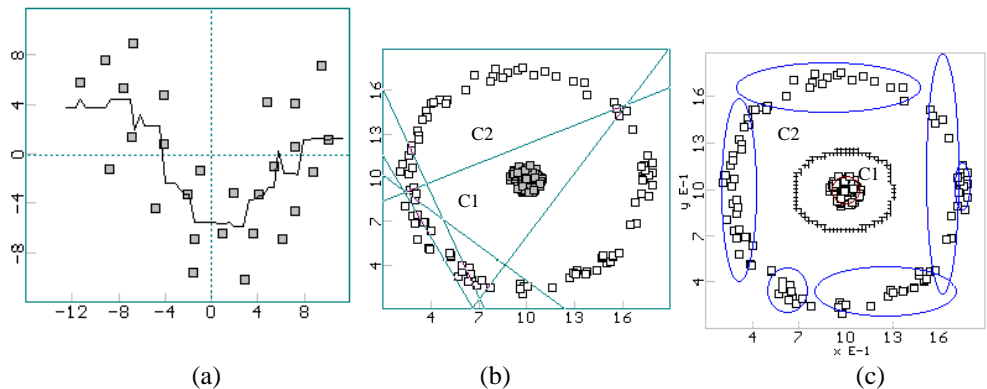

(a)              (b)              (c)

Figure 21. Alternative solutions in regression and classification (see text for details).

The nature of the solutions obtained by Backpropagation Multilayer Perceptrons (Rumelhart et al. 1986) in pattern classification is also illustrative. In general, each decision region can be geometrically expressed as the union of intersections of several half–spaces defined by the units in the first hidden layer. However, backprop networks often require very long learning times, many adjustable parameters and, what is worse, apparently simple distributions of patterns are hard to learn. For instance, the solution to the circle-ring classification problem in Fig. 21.b, obtained by a network with 6 hidden units requires hundreds of standard backprop epochs. The decision regions are not very satisfactory, even though the network has extra flexibility for this task (3 hidden units suffice to separate the training examples). Better solutions exist using all the resources in the network architecture, but backprop learning does not find them. In contrast, the solution obtained by the MFGN approach using 7 components (Fig. 21.c) requires a learning time orders of magnitude shorter than backprop optimization. All the components in the mixture contribute to synthesize reliable decision regions and acceptable solutions can be also obtained with a smaller number of components.

The proposed approach is closely related to a well-known family of approximation techniques which, essentially, distribute (using some kind of clustering or self-organizing algorithm) "detectors" over the relevant regions of the input space and then combine their responses for computing the desired outputs. This is the case of Radial Basis Functions (RBF) (Hertz et al.), the classification and regression trees proposed in (Breiman et al. 1984) and the topological maps used in (Cherkassky & Najafi 1992) to locate the "knots" required for piecewise linear regression.

A relevant methodology is proposed in (Jordan & Jacobs 1994, Peng et al. 1995), where the EM algorithm is used to learn hierarchical mixtures of experts in the form of linear rules in such a way that the desired posterior densities can be explicitly obtained. The properties of the EM algorithm are also satisfactorily used in (Ghahramani & Jordan 1994) to obtain unbiased approximations from missing data in a mixture-based framework similar to ours. Our framework extends this successful approach by exploiting the conjugate properties of the chosen universal approximation model: uncertain information of arbitrary complexity can be efficiently processed in the inference and learning stages.

The MFGN framework is appropriate for a moderated number of variables showing relatively complex dependencies. In contrast, Bayesian Networks satisfactorily addresses the case of a large number of variables with clear conditional independence relations. There are situations in which a certain subset of the variables in a Bayesian Network shows no explicit causal structure. This subdomain could be empirically modeled by a mixture model in order to be considered later as a composite node embedded in the whole network. If the subdomain can be conditionally isolated from the rest of variables through a set of communication nodes, the MFGN framework can be used to perform the required inferences.

Finally, mixture models are typically used for unsupervised classification: the examples are labeled with the index of the component with highest posterior probability. In fact, the MFGN framework explicitly finds clusters in the training set. Furthermore, continuous and symbolic attributes are allowed in the joint density, so the examples are clustered using an implicit probabilistic metric which automatically weighs all the (heterogeneous) attributes, even with missing and uncertain values. However, this method is effective only when the groups of interest have the same structure as the component densities. In order to simplify inference the mixture components have been selected with constraints (Gaussian, independent variables) which are not necessarily verified by the "natural" groups found in real applications.

A tentative possibility (inspired in a common heuristic clustering technique) consists of joining overlapping components (e.g., according to the Battachariya distance, a well-known bound on the Bayes error used in Statistical Pattern Recognition (Fukunaga 1990)). Unfortunately, our experiments indicate that the overlapping threshold is a free parameter that strongly determines the quality of the results. A universal threshold, independent of the application, does not seem to exist. In principle, clusters of arbitrary geometry may be discovered, but this cannot be easily automated. Therefore, other nonparametric cluster analysis methods (e.g. density valley seeking) are suggested for labeling complex groups.

## 5.4 Experimental Evaluation

The MFGN method has been evaluated on standard benchmarks from the Machine Learning database repository at the University of California, Irvine (Merz and Murphy 1996). It contains inductive learning problems which are representative of real world situations. We have experimented with the following databases: *Ionosphere*, *Pima Indians*, *Monk's Problems*, and *Horse Colic*, which illustrate different properties of the proposed methodology. In most cases MFGN has been compared to alternative learning methods with respect to the inference task considered of interest in each problem (typically, prediction of a specific attribute given the rest of them). We usually give the error rate over both the training and the test set to indicate the amount of overfitting obtained by the learning algorithms.

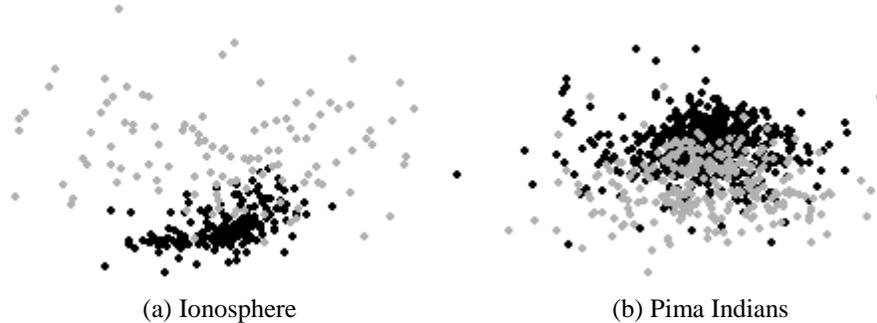

(a) Ionosphere                    (b) Pima Indians

Figure 22. Most discriminant 2D projections of two representative databases.

### 5.4.1 IONOSPHERE DATABASE

Two classes of radar returns from the ionosphere must be discriminated from vectors of 32 continuous attributes[7]. There are 351 examples, randomly partitioned into two disjoint subsets of approximately equal size for training and testing. The prevalence of the minoritary class (random prediction rate) is 36%. Figure 22.a and Table 7 show that this is a typical statistical pattern recognition problem, easily solvable by standard methods. The results suggest that the Bayes (optimum) error probability is around 5%.

| METHOD | error rate (training set) | $\hat{P}_E$ (test set) |
|---|---|---|
| Linear MSE (pseudoinverse) | .11 | .14 |
| 1-1 Nearest Neighbor | | .13 |
| 2-3 Nearest Neighbor | | .18 |
| Parzen Model | .05 | .08 |
| Backprop Multilayer Perceptron 2 hidden units | .00 | .08 |
| Support Vector Machine, RBF kernel, width 1, (105 s.v.) | | .05 |
| Support Vector Machine, RBF kernel, width 3, (35 s.v.) | | .09 |
| Support Vector Machine, polinomial kernel, order 2, (41 s.v.) | | .13 |
| Support Vector Machine, polinomial kernel, order 3, (45 s.v.) | | .17 |
| Support Vector Machine, polinomial kernel, order 4, (42 s.v.) | | .20 |
| Full covariance gaussian mixture, 1 component/class | .03 | .11 |
| Full covariance gaussian mixture, 2 component/class | .01 | .19 |
| Full covariance gaussian mixture, 3 component/class | .005 | .26 |
| MFGN 4 components (average) | .22±.15 | .21±.08 |
| MFGN 8 components (average) | .11±.06 | .13±.06 |
| MFGN 15 components (average) | .10±.05 | .13±.06 |
| MFGN, best result by cross-validation (8 components) | .07 | .06 |

Table 7. *Ionosphere* Database Results

---

[7] Originally the database contains 34 attributes. Two of them, meaningless or ill behaved, were eliminated.

In this problem, the plain MFGN method, without special heuristics in the learning stage, is comparable in average to the alternative methods. The best solution on the training set (cross-validation) is entirely satisfactory.

For the *Ionosphere* database we also present an exhaustive study of performance given varying proportions of missing values in the training and testing examples. A value of $x$ % means that in all training or test examples the value of each attribute is deleted with probability $x$. The basic experiment consists of learning a MFGN model with the prescribed number of components (4, 8 and 15) and computing the error rate on the training and test sets. Table 8 shows the mean value $\pm$ 2 standard deviations of the error rates obtained in 10 repetitions of the basic experiment in each configuration. Column $M$ contains the error rate of each configuration over its own training set. The training/test partition is kept fixed to analyze the variability of the solutions due to random initialization of the EM.

| LEARNING | | INFERENCE | | | |
|---|---|---|---|---|---|
| | $M$ | 0% | 10% | 25% | 50% |
| **4 COMP. - 0%** | $22 \pm 15$ | $21 \pm 8$ | $21 \pm 8$ | $22 \pm 8$ | $22 \pm 9$ |
| **8 COMP. - 0%** | $11 \pm 6$ | $13 \pm 6$ | $12 \pm 6$ | $13 \pm 5$ | $12 \pm 6$ |
| **15 COMP. - 0%** | $10 \pm 5$ | $13 \pm 6$ | $13 \pm 6$ | $13 \pm 5$ | $13 \pm 4$ |
| **4 COMP. - 10%** | $21 \pm 14$ | $23 \pm 11$ | $23 \pm 11$ | $23 \pm 11$ | $23 \pm 11$ |
| **8 COMP. - 10%** | $11 \pm 3$ | $13 \pm 5$ | $12 \pm 5$ | $13 \pm 5$ | $13 \pm 4$ |
| **15 COMP. - 10%** | $10 \pm 3$ | $12 \pm 6$ | $12 \pm 7$ | $12 \pm 6$ | $12 \pm 3$ |
| **4 COMP. - 25%** | $18 \pm 7$ | $19 \pm 5$ | $19 \pm 5$ | $18 \pm 6$ | $18 \pm 6$ |
| **8 COMP. - 25%** | $12 \pm 7$ | $14 \pm 10$ | $14 \pm 9$ | $15 \pm 8$ | $14 \pm 7$ |
| **15 COMP. - 25%** | $9 \pm 5$ | $12 \pm 9$ | $13 \pm 11$ | $13 \pm 9$ | $13 \pm 7$ |
| **4 COMP. - 50%** | $27 \pm 18$ | $26 \pm 15$ | $27 \pm 15$ | $27 \pm 14$ | $26 \pm 13$ |
| **8 COMP. - 50%** | $16 \pm 12$ | $21 \pm 15$ | $21 \pm 15$ | $21 \pm 13$ | $20 \pm 11$ |
| **15 COMP. - 50%** | $13 \pm 6$ | $26 \pm 17$ | $25 \pm 15$ | $25 \pm 14$ | $23 \pm 13$ |

Table 8. Evaluation of MFGN on *Ionosphere* Database given different proportions of missing data in the training and testing subsets.

As expected, the MFGN model is robust with respect to large proportions of missing values in the test patterns, and to moderated proportions of missing data in the training set. We have compared the above behavior with a standard algorithm for Decision Tree construction inspired in (Quinlan 1993), which is also able to support missing values[8]. Table 9 shows the error rates of the decision trees for the same experimental setting as in Table 8. This kind of Decision Tree obtains error rates that are better than the averages obtained by MFGN. However, MFGN's best solutions (selected by cross-validation) are better than the ones obtained by Decision Tree. Furthermore, Decision Tree performance degrades faster than MFGN, especially with respect to the proportion of missing values in the inference stage.

---

[8] Essentially, missing values are handled as follows. In the learning stage, when an attribute is selected, examples with missing values are sent to all the partitions with appropriate weights. In the inference stage, if a node asks for a missing value, it follows all the branches with appropriate weights and finally the outputs are combined.

| LEARNING | | INFERENCE | | | |
|---|---|---|---|---|---|
| | *M* | 0 % | 10 % | 25 % | 50 % |
| 0 % | 1 % | 9 % | 10 % | 11 % | 12 % |
| 10 % | 5 % | 14 % | 15 % | 19 % | 18 % |
| 25 % | 6 % | 15 % | 17 % | 17 % | 18 % |
| 50 % | 8 % | 17 % | 18 % | 18 % | 19 % |

Table 9. Evaluation of basic Decision Tree on *Ionosphere* Database given different proportions of missing data in the training and testing subsets.

### 5.4.2 PIMA INDIANS DATABASE

In this problem we must discriminate between two possible results of a diabetes test given to Pima Indians. There are 8 continuous attributes, and 768 examples, randomly partitioned into two disjoint subsets of equal size for training and testing. The prevalence of the minority class is 35%. The attribute vector has been normalized. Table 10 presents comparative results.

| METHOD | error rate (training set) | $\hat{P}_E$ (test set) |
|---|---|---|
| Linear MSE (pseudoinverse) | .22 | .23 |
| Oblique Decision Tree 8 decision nodes | .18 | .24 |
| 1-1 Nearest Neighbor | | .30 |
| 2-3 Nearest Neighbor | | .25 |
| Full covariance gaussian mixture, 1 component/class | .24 | .26 |
| Full covariance gaussian mixture, 2 component/class | .19 | .29 |
| Full covariance gaussian mixture, 3 component/class | .17 | .30 |
| Full covariance gaussian mixture, 4 component/class | .17 | .31 |
| Backprop Multilayer Perceptron 2 hidden units | .17 | .25 |
| Backprop Multilayer Perceptron 4 hidden units | .14 | .24 |
| Backprop Multilayer Perceptron 8 hidden units | .05 | .29 |
| Support Vector Machine, RBF kernel, width 1 (297 s.v.) | | .30 |
| Support Vector Machine, RBF kernel, width 3 (176 s.v.) | | .35 |
| Support Vector Machine, polynomial kernel, order 4 (138 s.v.) | | .36 |
| Support Vector Machine, polynomial kernel, order 5 (131 s.v.) | | .34 |
| MFGN 4 components | .28 | .35 |
| MFGN 6 components | .25 | .32 |
| MFGN 8 components | .29 | .35 |

Table 10. *Pima Indians* Database Results

Despite of low dimensionality and large number of examples, this classification problem is hard (see Figure 22.b). Even sophisticated learners such as backpropagation networks, decision trees or support vector machines, which are able to store a reasonable proportion of the training set, do not achieve significant generalization. MFGN shows a similar behavior, although it is slightly less prone to overfitting (the error rate on the training set is not misleading).

### 5.4.3 HORSE COLIC DATABASE

This database contains a classification task from a heterogeneous attribute vector including symbolic, discrete and continuous variables, with 30% missing values. It illustrates the problem of feature selection in the context of joint modeling, mentioned in Section 5.1. Table 11 shows the error rates obtained by MFGN using different attribute subsets[9]. To take advantage of its general inference properties, the MFGN model must be applied to the attribute subset of interest. If the inference task is fixed and the number of attributes is very large, alternative methods should be used.

| METHOD | $\hat{P}_E$ (distribution, 10 initializations) | $\hat{P}_E$ (best) |
|---|---|---|
| *6 Selected attributes* | | |
| MFGN 2 components | .32±.00 | .32 |
| MFGN 3 components | .20±.05 | .18 |
| MFGN 4 components | .19±.02 | .18 |
| MFGN 5 components | .20±.02 | .18 |
| MFGN 6 components | .19±.03 | .16 |
| MFGN 7 components | .20±.04 | .18 |
| MFGN 10 components | .22±.04 | .18 |
| MFGN 12 components | .19±.03 | .16 |
| MFGN 15 components | .19±.04 | .15 |
| | | |
| *8 Selected attributes* | | |
| MFGN 4 components | .22±.01 | .21 |
| MFGN 6 components | .21±.02 | .19 |
| MFGN 8 components | .21±.03 | .18 |
| MFGN 10 components | .23±.02 | .18 |
| MFGN 12 components | .21±.02 | .18 |
| MFGN 15 components | .21±.02 | .18 |
| | | |
| *23 Selected attributes* | | |
| MFGN 6 components | .28±.02 | .25 |
| MFGN 8 components | .29±.03 | .25 |
| MFGN 10 components | .34±.08 | .25 |
| MFGN 15 components | .34±.06 | .28 |

Table 11. *Horse Colic* Database Results (random rate = .5)

### 5.4.4 MONK'S PROBLEMS

The Monk's problems are three concept learning tasks from 6 symbolic attributes, widely used as benchmarks for inductive learning algorithms (Thrun et al. 1991). As seen in Table 12, MFGN fails on MONK1 (where acceptable generalization is not obtained) and MONK2 (where the training examples cannot even be stored). In contrast, MFGN correctly solves MONK3. This behavior is related to the fact that the MONK's problems are based on deterministic or abstract concepts which may lack the kind of geometric regularities in the attribute space required by probabilistic models[10].

---

[9] Features were individually selected using a simple discrimination index related to the Kolmogorov-Smirnov statistic (Ruiz 1995).

[10] A typical example is the *parity* problem: acceptable off-training-set generalization cannot be achieved if the inductive bias of the learning machine is biased towards "smooth" solutions.

Fig. 23 shows the most discriminant 2D projections of the datasets and illustrates the fact that MONK2 cannot be easily captured by statistic techniques. In this benchmark, MFGN performance is similar to that of other popular probabilistic methods (Thrun et al. 1991).

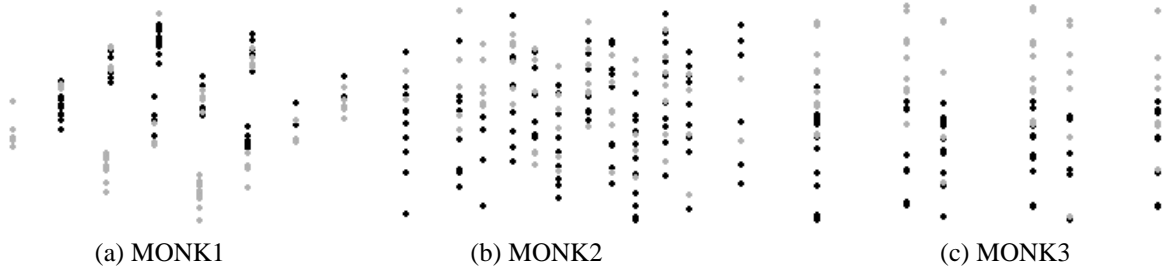

(a) MONK1          (b) MONK2          (c) MONK3

Figure 23. Most Discriminant 2D Projections of the Monk's Datasets.

| METHOD | error rate *(training set)* | $\hat{P}_E$ *(test set)* |
|---|---|---|
| | | |
| **MONK1 (random rate = .5)** | | |
| Linear MSE (pseudoinverse) | .29 | .34 |
| 1-1 Nearest Neighbor | | .17 |
| Support Vector Machine, RBF kernel, width 1 (78 s.v.) | | .08 |
| Cascade Correlation | | 0 |
| MFGN 4 components | .06 | .40 |
| MFGN 8 components | .00 | .33 |
| | | |
| **MONK2 (random rate $\cong$ .4)** | | |
| Linear MSE (pseudoinverse) | .40 | .37 |
| 1-1 Nearest Neighbor | | .19 |
| Support Vector Machine, RBF kernel, width 1 (117 s.v.) | | .20 |
| Cascade Correlation | | 0 |
| MFGN 4 components | .31 | .38 |
| MFGN 8 components | .26 | .44 |
| MFGN 15 components | .14 | .50 |
| | | |
| **MONK3 (random rate $\cong$ .5)** | | |
| Linear MSE (pseudoinverse) | .19 | .19 |
| 1-1 Nearest Neighbor | | .18 |
| Support Vector Machine, RBF kernel, width 1 (69 s.v.) | | .08 |
| Cascade Correlation | | .03 |
| MFGN 2 components | .07 | .03 |
| MFGN 4 components | .04 | .03 |
| MFGN 8 components | .03 | .08 |

Table 12. *Monk's Problems* Results

### 5.4.5 COMMENTS

The above experiments demonstrate that the MFGN model is able to obtain acceptable results on many real world applications. In particular, the error rates obtained in standard classification tasks are comparable to those obtained by other popular learners. Additionally, MFGN is able to perform inferences over any other attribute given uncertain or partial information, which is not possible for most of the alternative methods. This property makes MFGN a very attractive alternative for many inference problems such as the one illustrated in Example 16. The experiments have also contributed to characterize the kind of problems for which the MFGN model is best suited. Essentially, the relationship among attributes must be of a true probabilistic nature, and the attribute vector must be of a moderated size containing "relevant" variables. A previous feature selection / accommodation stage is recommended in certain applications.

## 6. Conclusions

We have developed an efficient methodology for probabilistic inference and learning from uncertain information. Under the proposed MFGN framework, the joint probability density function of the attributes and the likelihood function of the available information are approximated by Mixtures of Factorized Generalized Normals. This mathematical structure allows efficient computation, without numerical integration, of posterior densities and expectations of the desired variables given events of arbitrary "geometry". An extended version of the EM learning algorithm has been developed to estimate the parameters of the required mixture models from uncertain training examples. Different paradigms as pattern recognition, regression or pattern completion are subsumed under a common framework.

A comprehensive collection of examples illustrates the methodology, which has been critically compared with alternative techniques. The Extended EM algorithm is able to learn satisfactory domain models from a reasonable number of examples with uncertain values, taking into account the explicit likelihood functions of the available information. Results are satisfactory whenever the sample size is large in relation to the amount of (known) degradation of the training set. The experiments also characterized the kind of situations that the model manages better: Domains described by a moderate number of heterogeneous attributes with complex probabilistic dependences, problems in which the output variables are not necessarily known in the learning stage (i.e. pattern completion), and, finally, problems in which an explicit management of uncertainty is needed, either in the learning or in the inference stage (or even in both). The MFGN framework has obtained a very favorable trade-off between useful features and model complexity in the solutions to different applications and benchmarks.

Future developments of our work include improving the learning stage with some heuristic steps that are combined with the standard $E$ and $M$ steps to control the adequacy of the acquired models. Additional studies are required on validation tests, generalization, scalability, robustness and data preprocessing. The essential idea of working with explicit likelihood functions will be incorporated into the Parzen approximation scheme and we are also interested in more expressive model structures such as mixtures of factor analyzers, principal component analyzers or linear experts. Finally, the methodology can be developed in a pure Bayesian framework or subsumed under the Dempster-Shafer Evidence Theory.

## Acknowledgments

The authors would like to thank the anonymous reviewers for their careful reading and helpful suggestions. This work has been supported by the Spanish CICYT grants TIC95-1019, TIC97-1343-C02-02 and TIC97-0897-C04-03.

## References

Berger, J., (1985). *Statistical Decision Theory and Bayesian Analysis*. Springer-Verlag.

Bernardo, J.M., Smith, A.F.M. (1994). *Bayesian Theory*. Wiley.

Bouckaert, R.R. (1994). "Properties of Bayesian Belief Network Learning Algorithms". *Proceedings of Uncertainty in AI*, pp. 102-109.

Breiman, L., Friedman, J.H., Olshen, R.A., and Stone, C.J. (1984). *Classification and Regression Trees*. Wadsworth International Group, Belmont, CA.

Chang, K. & Fung, R. (1995). "Symbolic Probabilistic Inference with Both Discrete and Continuous Variables". *IEEE Tran. on Systems, Man, and Cybernetics*, Vol. 25, No. 6, june, pp. 910-916.

Cherkassky, V. and Lari-Najafi, H. (1992). "Nonparametric Regression Analyisis Using Self-Organizing Topological Maps" in H. Wechsler (ed.), *Neural Networks for Perception. Vol.2, Computation, Learning and Architectures*, San Diego: Academic Press.

Cohn, D.A., Ghahramani, Z. & Jordan, M.I. (1996). "Active Learning with Statistical Models". *Journal of Artificial Intelligence Research* 4, pp. 129-145.

Dalal, S.R. & Hall, W.J. (1983). "Approximating Priors by Mixtures of Natural Conjugate Priors". *J. R. Statist. Soc. B*, Vol. 45, No. 2, pp. 278-286.

De Soete, G. (1993). "Using Latent Class Analysis in Categorization Research" in I. V. Mechelen, J. Hampton, R.S. Michalski, P. Theuns (eds.), *Categories and Concepts: Theoretical Views and Inductive Data Analysis*, San Diego: Academic Press.

Dempster, A.P., Laird, N.M., Rubin, D.B., (1977). "Maximum Likelihood Estimation from Incomplete Data via the EM Algorithm". *Journal of the Royal Statistical Society*, Series B, Vol. 39: pp. 1-38.

Duda, R.O. and Hart, P.E. (1973). *Pattern Classification and Scene Analysis*. John Wiley & Sons.

Fan, C.M., Namazi, N.M. and Penafiel, P.B. (1996). "A New Image Motion Estimation Algorithm Based on the EM Technique". *IEEE Transactions on Pattern Analisys and Machine Intelligence*, Vol.18, No.3, March, pp. 348-352.

Fukunaga, K. (1990). *Introduction to Statistical Pattern Recognition*. Academic Press.

Ghahramani, Z. and Jordan, M.I. (1994) "Supervised learning from Incomplete data via an EM approach" In Cowan, J.D., Tesauro, G., and Alspector, J. (eds.). *Advances in Neural Information Processing Systems 6*. Morgan Kauffman

Ghahramani, Z. and Hinton, G.E. (1996) "The EM algorithm for mixtures of factor analyzers". Tech. Rep. Univ. Toronto. CRG-TR-96-1.

Heckerman, D. & Wellman, M.P. (1995). "Bayesian Networks". *Communications of the ACM*, Vol. 28, No.3, pp. 27-30, March.

Hertz, J., Krogh, A., Palmer, R.G., (1991). *Introduction to the Theory of Neural Computation*. Addison Wesley.

Hinton, G.E., Dayan, P. and Revow, M. (1997). "Modeling the manifold of images of handwritten digits". *IEEE T. on Neural Networks* 8, pp. 65-74.

Hornik, K., Stinchcombe, M., White, H., (1989). "Multilayer FeedForward Networks are Universal Approximators". *Neural Networks*, No.2.

Hutchinson, A. (1994). *Algorithmic Learning*. New York: Oxford Univ. Press.

Izenman, A.J. (1991). "Recent Developments in Nonparametric Density Estimation". *J. Amer. Statist. Assoc.* Vol. 86, No. 413, pp. 205-224.

Jordan, M.I., Jacobs, R.A., (1994). "Hierarchical Mixtures of Experts and the EM Algorithm". *Neural Computation*, 6, pp. 181-214.

Kohonen, T., (1989). *Self-Organization and Associative Memory*. Springer-Verlag.

Lauritzen, S.L. & Spiegelhalter, D. J. (1988). "Local Computations with Probabilities on Graphical Structures and their Application to Expert Systems". *J. R. Statist. Soc. B.* 50, No. 2, pp. 157-224.

Li, M. and Vitànyi, P. (1993). *An Introduction to Kolmogorov Complexity and Its Applications*. New York: Springer-Verlag.

McLachlan, G.J., Basford, K.E., (1988). *Mixture Models*. New York: Marcel Dekker.

McLachlan, G.J. and Krishnan, T. (1997). *The EM Algorithm and Extensions*. John Wiley and Sons.

Michalski, R.S., Carbonell, J. and Mitchell, T.M., eds. (1983). *Machine Learning: An Artificial Intelligence approach*. Palo Alto, CA: Tioga Press. Also reprinted by Morgan Kaufmann (Los Altos, CA).

Michalski, R.S., Carbonell, J. and Mitchell, T.M., eds. (1986). *Machine Learning: An Artificial Intelligence approach, Vol. II*. Los Altos, CA: Morgan Kaufmann.

Mohgaddam, B. and Pentland, A. (1997). "Probabilistic Visual Learning for Object Representation". *IEEE T PAMI,* Vol. 19, No.7, 710. pp. 696-.

Merz, C.J. and Murphy, P.M. (1996). *UCI Repository of machine learning databases*. [http://www.ics.uci.edu/~mlearn/MLRepository.html]. Irvine, CA: University of California, Department of Information and Computer Science.

Palm, H.C. (1994). "New method for generating statistical classifiers assuming linear mixtures of Gaussian densities", *Proceedings 12th IAPR International Conference on Pattern Recognition* (Jerusalem, October 9-13, 1994), vol.2, IEEE, Piscataway, NJ, USA,

Papoulis, A., (1991). *Probability, Random Variables and Stochastic Processes*. MCGraw-Hill.

Pearl, J., (1988). *Probabilistic reasoning in intelligent systems: Networks of plausible inference*. Morgan Kaufmann.

Peng, F., Jacobs, R.A., Tanner, M.A. (1995). "Bayesian Inference in Mixtures-of-Experts and Hierchical Mixtures-of-Experts Models With an Application to Speech Recognition". Accepted in the *Journal of the American Statistical Association*.

Priebe, C.E., Marchette, D.J., (1991). "Adaptive mixtures: recursive nonparametric pattern recognition". *Pattern Recognition*, V24 N12, pp. 1197-1209.

Pudil, P., Novovicova, J., Choakjarernwanit, N., Kittler, J. (1995). "Feature Selection Based on the Approximation of Class Densities by Finite Mixtures of Special Type". *Pattern Recognition* Vol.28 No.9 pp. 1389-1398.

Quinlan, J.R., (1993). *C4.5: Programs for Machine Learning*. San Mateo, CA: Morgan Kaufmann.

Redner, R.A., Walker, H.F., (1984). "Mixture densities, maximum likelihood estimation and the EM algorithm". *SIAM Review*, Vol. 26, pp. 195-239.

Rojas, R. (1996). "A Short Proof of the Posterior Probability Property of Classifier Neural Networks". *Neural Computation* Vol.8 Issue 1, January.

Rumelhart, D.E., Hinton, G. E. and Williams, R. (1986). "Learning Internal Representations by Error Propagation" in Rumelhart, McClelland & the PDP Group (1986), pp. 319-362.

Rumelhart, D.E., McClelland, J.L. & the PDP Research Group. (1986). *Parallel Distributed Processing: Explorations in the Microstructure of Cognition,* vol. 1, *Foundations.* Cambrigde MA, Bradford Books/MIT Press.

Ruiz, A. (1995). "A nonparametric bound for the Bayes Error". *Pattern Recognition*, Vol. 28, No. 6, pp. 921-930.

Ruiz, A., López-de-Teruel, P.E. and Garrido, M.C. (1998). "Kernel Density Estimation from Indirect Observations". In preparation.

Sung, K.-K. and Poggio, T. (1998),. "Example Based Learning for View-Based Human Face Detection". *IEEE Trans. Pattern Analyisis and Machine Intelligence.* Vol.20, N.1. January, pp. 39-51.

Tanner, M.A. (1996). *Tools for statistical inference. (3$^{rd}$ ed.).* Springer.

Thrun, S. et al (1991). "The MONK's Problems. A performance Comparison of Different Learning Algorithms". *Technical Report* CMU-CS-91-197.

Titterington, D.M., A.F.M. Smith and U.E. Makov (1985). *Statistical Analysis of Finite Mixture Distributions*, Wiley, New York.

Traven, H.G.C., (1991). "A Neural Network Approach to Statistical Pattern Classification by "Semiparametric" Estimation of Prob. Den. Func.". *IEEE T Neural Networks*, V2 N3.

Valiant, L.G. (1993). "A View of Computational Learning Theory" in Meyrowitz and Chipman, eds. (1993). *Foundations of Knowledge Acquisition: Machine Learning.* Kluwer Acad. Pub.

Valiveti, R.S., Oommen, B.J., (1992). "On using the chi-squared metric for determining stochastic dependence". *Pattern Recognition*, V25 N11 pp. 1389-1400.

Vapnik, V.N. (1982). *Learning Dependencies based on Empirical Data.* Springer, New York.

Vapnik, V.N. (1995). *The Nature of Statistical Learning Theory*. Springer, New York.

Wan, E.A., (1990). "Neural Networks Classification: A Bayesian Interpretation". *IEEE Trans. on Neural Networks*, V1 N4.

Weiss, Y. and Adelson E.H. (1995). "Perceptually organized EM: A framework for motion segmentation that combines information about form and motion". *TR MIT MLPCS TR* 315.

Wilson, D.R. and Martinez, T.R. (1997). "Improved Heterogeneous Distance Functions". *JAIR*, V6, pp. 1-34.

Wolpert, D.H., ed. (1994a). *The Mathematics of Generalization*. Proc. SFI/CLNS workshop on Formal Approaches to Supervised Learning.

Xu, L. & Jordan, M.I. (1996). "On the Convergence Properties of the EM Algorithm for Gaussian Mixtures". *Neural Computation* Vol.8 Issue 1, January.

You, Y.L. and Kaveh, M. (1996). "A regularization approach to joint blur identification and image restoration". *IEEE T on Image Processing,* vol 5.no.3, pp. 416-428.